\documentclass[conference]{IEEEtran}
\IEEEoverridecommandlockouts
\usepackage{cite}
\usepackage{amsmath,amssymb,amsfonts}
\usepackage{algorithmic}
\usepackage{graphicx}
\usepackage{textcomp}
\usepackage{booktabs}
\usepackage{xcolor}
\def\BibTeX{{\rm B\kern-.05em{\sc i\kern-.025em b}\kern-.08em
    T\kern-.1667em\lower.7ex\hbox{E}\kern-.125emX}}
\usepackage{enumitem}
    
\usepackage{siunitx}
\usepackage{tikz}
\usepackage{blindtext}
\usepackage{hyperref}
\usepackage{multirow}

\usepackage{dblfloatfix}

\usepackage{fancyhdr}

\newcommand{\uproman}[1]{\uppercase\expandafter{\romannumeral#1}}    
    
\newcommand{\discolorlinks}[1]{{\hypersetup{hidelinks}#1}}
    
\begin{document}

\title{Efficient Multi-Task RGB-D Scene Analysis for Indoor Environments
\thanks{This work has received funding from the German Federal Ministry of Education and Research (BMBF) to the project MORPHIA~(grant agreement no. 16SV8426) and from Carl-Zeiss-Stiftung to the project CO-HUMANICS.}
}

\author{\IEEEauthorblockN{Daniel Seichter, Söhnke Benedikt Fischedick, Mona Köhler, and Horst-Michael Groß}
\IEEEauthorblockA{\textit{Ilmenau University of Technology, Neuroinformatics and Cognitive Robotics Lab} \\
98684 Ilmenau, Germany \\
\small\tt daniel.seichter@tu-ilmenau.de, ORCID:
\discolorlinks{\href{https://orcid.org/0000-0002-3828-2926}{0000-0002-3828-2926}}}
}

\maketitle

\newboolean{isarxiv}
\setboolean{isarxiv}{true}
\ifthenelse{\boolean{isarxiv}}{%
    \renewcommand{\headrulewidth}{0pt}
    \fancypagestyle{fancyfirstpage}{%
        \fancyhf{}%
        \fancyhead[C]{%
            \scriptsize%
            \textcolor{gray}{%
                © 2022 IEEE.  Personal use of this material is permitted.  Permission from IEEE must be obtained for all other uses, in any current or future media, including reprinting/republishing this material for advertising or promotional purposes, creating new collective works, for resale or redistribution to servers or lists, or reuse of any copyrighted component of this work in other works.
            }%
        }
        \fancyfoot[C]{%
            \footnotesize%
            \textcolor{gray}{\thepage}%
        }
    }
    \fancypagestyle{fancypage}{%
        \fancyhf{}%
        \fancyfoot[C]{%
            \footnotesize%
            \textcolor{gray}{\thepage}%
        }        
    }    
    \thispagestyle{fancyfirstpage}
    \pagestyle{fancypage}
}{%
    \thispagestyle{empty}%
    \pagestyle{empty}%
}%

\begin{abstract}
Semantic scene understanding is essential for mobile agents acting in various environments.
Although semantic segmentation already provides a lot of information, details about individual objects as well as the general scene are missing but required for many real-world applications.
However, solving multiple tasks separately is expensive and cannot be accomplished in real time given limited computing and battery capabilities on a mobile platform. 
In this paper, we propose an efficient multi-task approach for RGB-D scene analysis~(EMSANet) that simultaneously performs semantic and instance segmentation~(panoptic segmentation), instance orientation estimation, and scene classification.
We show that all tasks can be accomplished using a single neural network in real time on a mobile platform without diminishing performance\mbox{~--~}by contrast, the individual tasks are able to benefit from each other.
In order to evaluate our multi-task approach, we extend the annotations of the common RGB-D indoor datasets NYUv2 and SUNRGB-D for instance segmentation and orientation estimation. 
To the best of our knowledge, we are the first to provide results in such a comprehensive multi-task setting for indoor scene analysis on NYUv2 and SUNRGB-D. 

\end{abstract}

\begin{IEEEkeywords}
Multi-task learning, 
orientation estimation, 
panoptic segmentation, 
scene classification, 
semantic segmentation,
NYUv2, SUNRGB-D
\end{IEEEkeywords}

\section{Introduction}
\label{sec:introduction}

\begin{figure}[!t]
    \vspace{0.5mm}
	\centering
	\includegraphics[width=\columnwidth]{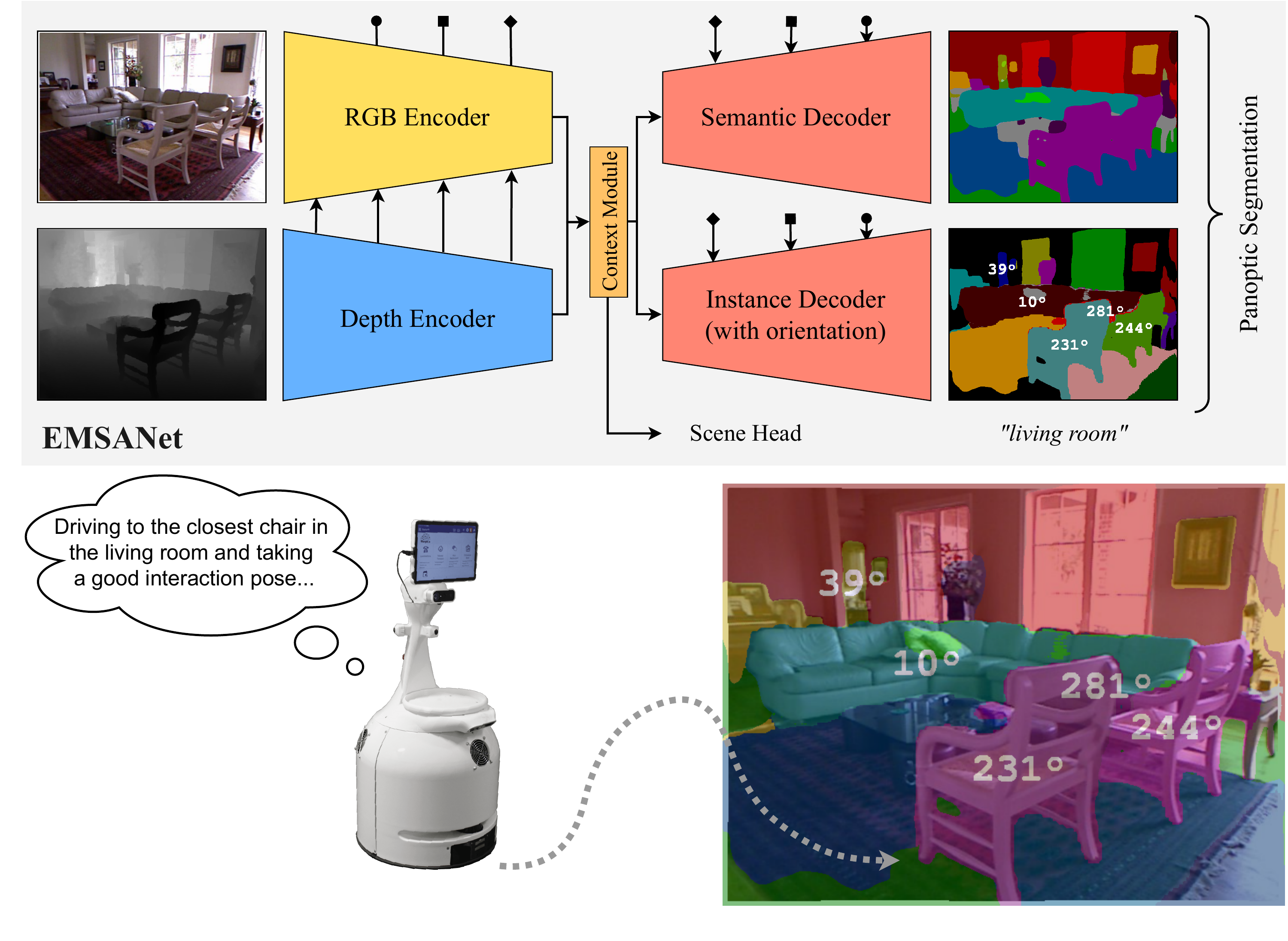}
	\vspace{-9mm}
	\caption{Prediction of our proposed Efficient Multi-task Scene Analysis Network~(EMSANet) that simultaneously performs panoptic segmentation, orientation estimation, and scene classification. With 24.5 FPS on an NVIDIA Jetson AGX Xavier it is well suited for mobile robotic applications. See Fig.~\ref{fig:iou-radar-chart} for semantic label colors. %
	Color variations indicate individual instances.
	}
	\label{fig:eyecatcher}
	\vspace{-4mm}
\end{figure}

In computer vision, semantic scene understanding is often equated with semantic segmentation as it enables gaining precise knowledge about the structure of a scene by assigning a semantic label to each pixel of an image.
However, this kind of knowledge is not sufficient for the agents in our ongoing research projects \href{http://morphia-projekt.de/}{MORPHIA} and \href{http://co-humanics.de/}{CO-HUMANICS} that require operating autonomously in their environments.
Imagine a mobile robot that is supposed to navigate to a semantic entity, e.g., a specific chair within a group of chairs in the living room, as shown in Fig.~\ref{fig:eyecatcher}. 
Performing such a high-level task requires a much broader understanding of the scene.
First, even with a semantic map of the environment~\cite{semanticmapping2022icra}, the robot still needs to know which part of its environment belongs to the living room.
Subsequently, it needs to be able to distinguish individual instances of the same semantic class, and, finally, for approaching the chair from the right direction, its orientation is required.

In this paper, we present an approach called Efficient Multi-task Scene Analysis Network~(EMSANet) for tackling all the aforementioned challenges in order to accomplish such a high-level task.
Our approach performs scene classification, semantic and instance segmentation (panoptic segmentation), as well as instance orientation estimation.
However, given limited computing and battery resources on a mobile platform, solving all these tasks separately is expensive and cannot be accomplished in real time.
Therefore, we design our approach to solve all aforementioned tasks using a single efficient multi-task network.
Our approach extends ESANet~\cite{esanet2021icra}, an efficient approach for semantic segmentation, by adding additional heads for tackling panoptic segmentation, instance orientation estimation, and scene classification.
ESANet processes both RGB and depth data as input. 
As shown in~\cite{esanet2021icra}, especially for indoor environments, depth data provide complementary geometric information that help analyzing cluttered indoor scenes.
In this paper, we show that this also holds true for panoptic segmentation, instance orientation estimation, and scene classification.
Thus, our approach also relies on both RGB and depth data.

Training such a multi-task approach requires comprehensive datasets.
However, to the best of our knowledge, there is no real-world RGB-D indoor dataset encompassing ground-truth annotations for all aforementioned tasks.
Therefore, we enrich the existing datasets NYUv2~\cite{NYUv2-eccv2012} and SUNRGB-D~\cite{SUNRGBD-cvpr2015} with additional annotations for instance segmentation and instance orientation estimation.
With this data at hand, we first train single-task baselines and subsequently combine multiple tasks in several multi-task settings.
We experimentally show that all tasks can be solved using a single neural network in real time without diminishing performance -- by contrast, the individual tasks are able to boost each other.
Our full multi-task approach reaches 24.5 FPS on the mobile platform NVIDIA Jetson AGX Xavier, while achieving state-of-the-art performance. 
Thus, it is well suited for real-world applications on mobile platforms.

\noindent In summary, our main contributions are:
\begin{itemize}[leftmargin=5mm]
    \item an efficient RGB-D multi-task approach for panoptic segmentation, scene classification, and instance orientation estimation~(EMSANet) including a novel encoding for instance orientations
    \item enriched annotations for NYUv2 and SUNRGB-D%
    \item detailed experiments regarding performance of each task in single- and multi-task settings as well as corresponding inference throughputs on an NVIDIA Jetson AGX Xavier.
\end{itemize}

Our code, the additional annotations for NYUv2 and SUNRGB-D as well as the trained models are publicly available at: {\small\href{https://github.com/TUI-NICR/EMSANet}{\texttt{\url{https://github.com/TUI-NICR/EMSANet}}}}.

\section{Related Work}
\label{sec:related_work}
In the following, we briefly summarize related work for each task. 
Moreover, we give some insights on combining tasks in multi-task settings.

\subsection{Semantic Segmentation}
\label{sec:related_work:semantic_segmentation}
Architectures for semantic segmentation typically follow an encoder-decoder design to accomplish dense pixel-wise predictions.
Well-known approaches such as PSPNet~\cite{PSPNet-cvpr2017} or the DeepLab series~\cite{DeepLab-iclr2015, DeepLabv3-arxiv2017, DeepLabv3plus-eccv-2018} achieve good results but cannot be executed in real time on mobile platforms due to their low downsampling of intermediate feature representation.
Thus, another line of research emerged, focusing on low inference time while still keeping high performance.
For example, ERFNet~\cite{ERFNet-its2018} introduces a more efficient block by spatially factorizing the expensive $3{\times}3$ convolution into a $3{\times}1$ and a $1{\times}3$ convolution and, thus, reduces computational effort.
By contrast, SwiftNet~\cite{SwiftNet-cvpr2019} simply uses a pretrained ResNet18~\cite{ResNet-cvpr2016} as encoder with early and high downsampling, resulting in low inference time but still good performance as well.

While the aforementioned approaches only process RGB data, especially for indoor applications, others~\cite{FuseNet-accv2016, ACNet-icip2019, SSMA-ijcv2019, RDFNet-iccv2017, ShapeConv-cvpr2021} also incorporate depth data as they provide complementary geometric information that help analyzing cluttered scenes.
Most approaches use two encoders for processing RGB and depth data~(RGB-D) first separately and fuse the resulting features later in the network.
However, almost all RGB-D approaches use deep and complex network architectures and do not focus on fast inference.
By contrast, our recently published ESANet~\cite{esanet2021icra} combines the merits of efficient and \mbox{RGB-D} semantic segmentation. 
It utilizes a carefully designed architecture featuring a dual-branch RGB-D ResNet-based encoder with high downsampling and spatially factorized convolutions enabling fast inference.
Our experiments in~\cite{esanet2021icra} reveal that processing both RGB and depth data with shallow backbones is superior to utilizing only RGB data and a deeper backbone.

Therefore, our approach follows ESANet and extends its architecture with additional heads tackling the remaining tasks.

\subsection{Panoptic Segmentation}
\label{sec:related_work:panoptic_segmentation}
Panoptic segmentation~\cite{PanopticSegmentation-cvpr2019} was introduced to unify semantic segmentation~(assigning a class label to each pixel) and instance segmentation~(assigning a unique id to pixels of the same instance) in a single task.
In panoptic segmentation, semantic classes for countable objects are regarded as thing classes and represent foreground.
Background classes, such as wall or floor -- known as stuff classes -- do not require instances.
Thus, all associated pixels have the same instance id.
Approaches for panoptic segmentation can be categorized in top-down, bottom-up, and end-to-end approaches.
Top-down approaches typically extend two-stage instance segmentation approaches such as Mask R-CNN~\cite{MaskRCNN-iccv2017} with an additional decoder for semantic segmentation~\cite{PanopticFPN-cvpr2019, UPSNet-cvpr2019}.
Although top-down approaches typically achieve superior performance, they have several major drawbacks:~As instance segmentation approaches can output overlapping instance masks, further logic is required to resolve these issues in order to merge instance and semantic segmentation without contradictions.
Moreover, they require complex training and inference pipelines, making them less suitable for mobile applications.
On the other hand, bottom-up approaches extend encoder-decoder-based architectures for semantic segmentation and separate thing classes into instances by grouping their pixels into clusters~\cite{DeeperLab-arxiv2019, SSAP-iccv2019, PanopticDeeplab-cvpr2020}.
As bottom-up approaches neither require region proposals, estimating multiple masks independently, nor further refinement steps, their pipelines for training and inference are much simpler compared to top-down approaches.
However, until Panoptic DeepLab~\cite{PanopticDeeplab-cvpr2020} bottom-up approaches could not compete with top-down approaches in terms of panoptic quality.
Nevertheless, both top-down and bottom-up approaches require additional logic for merging instance and semantic segmentation. 
The recently proposed MaX-DeepLab~\cite{Max-Deeplab-cvpr2021} follows another approach based on a novel dual-path transformer architecture~\cite{Transformer-neurips2017} and attempts to directly predict the panoptic segmentation using an end-to-end pipeline.
However, research for this kind of approaches currently focuses on establishing new architectures and not on fast and efficient inference.

Unlike for semantic segmentation, there are only a few approaches targeting efficiency~\cite{EfficientPTS-ijcv2021, LPSNet-cvpr2021, EPSNet-accv2020, RealTime-Panoptic-cvpr2020, FastPanoptic-ral2020}. 
However, their target hardware is different as they only report inference times on high-end GPUs. 
Execution on mobile platforms, such as an NVIDIA Jetson AGX Xavier, is expected to be much slower.

Our approach follows the bottom-up idea as it is straightforward to be integrated into ESANet and expected to enable faster inference on mobile platforms.

\subsection{Orientation Estimation}
\label{sec:related_work:orientation_estimation}
Orientation estimation is often done along with 3D bounding box detection~\cite{3DBBox-Estimation-cvpr2017, monocular3DObjectDetection-cvpr2019, ObjectsAreDifferent3D-cvpr2021} and deeply integrated into such architectures. 
Adapting these detectors to also accomplish dense predictions would require fundamental changes and, thus, is not suitable for our application.
Another field of research strongly related to orientation estimation is person perception~\cite{OpenPose-cvpr2017,Biternion-gcpr2015,DeepOrientation-iros2019,RealtimePersonOrientationEstimation-ecmr2019,MTPersonPerception-iros2020}.
Besides estimating a person's orientation inherently using its skeleton~\cite{OpenPose-cvpr2017}, there are also approaches directly estimating the orientation from patches~\cite{Biternion-gcpr2015, RealtimePersonOrientationEstimation-ecmr2019, DeepOrientation-iros2019,MTPersonPerception-iros2020}.
This can be performed using either classification or regression. 
However, as shown in~\cite{Biternion-gcpr2015}, classification adds further discretization inaccuracy and does not account well for periodicity. 
Therefore, approaches such as~\cite{DeepOrientation-iros2019, Biternion-gcpr2015} rely on regression and estimate the angle through its sine and cosine parts, which is often called Biternion encoding~\cite{Biternion-gcpr2015}.
The same authors also proposed to use the von Mises loss function~\cite{Biternion-gcpr2015} instead of L1 or MSE loss as it further improves accounting periodicity and avoiding discontinuities.

Our approach follows the latter idea and formulates orientation estimation as regression.
However, instead of using a patch-based approach, we propose a novel way to accomplish dense orientation estimation.

\subsection{Scene Classification}
\label{sec:related_work:scene_classification}
Scene classification, i.e., assigning a scene label such as kitchen or living room to an input image, is similar to other classification tasks such as the ImageNet-Challenge~\cite{ImageNet-ijcv2015}.
Thus, well known architectures~\cite{ResNet-cvpr2016, SENet-cvpr2018, MobileNetv2-cvpr2018, EfficientNet-icml2019} can be used.

\subsection{Multi-task Learning}
\label{sec:related_work:multitask}
Multi-task learning refers to learning multiple tasks simultaneously in a single neural network.
As these tasks commonly share at least some network parameters, inference is faster compared to using an independent network for each task.
Moreover, in~\cite{MTL-which-tasks-together-icml2020} it is shown, that dense prediction tasks may benefit from another when trained together. 
Especially early network layers are known to learn common features and, thus, can be shared among multiple tasks~-- in literature this is referred to as hard-parameter sharing~\cite{MTL-survey-tpami-2021}.
Some approaches~\cite{MTINet-eccv2020, JTRL-eccv2018} also exchange information in the task-specific heads, which is called soft-parameter sharing.
However, when utilizing soft-parameter-shared task heads, these tasks cannot be decoupled anymore.
This means that the whole network needs to be applied during inference, even though only a single task may be of interest.
Therefore, our approach uses a hard-parameter shared \mbox{RGB-D} encoder and independent task-specific heads, not sharing any network parameters or information.
We show that semantic and instance segmentation, instance orientation estimation as well as scene classification benefit from such a multi-task setting.
\section{Efficient Multi-task RGB-D Scene Analysis}
\label{sec:main}

\begin{figure*}[!t]
    \centering%
    \vspace{1mm}%
        \resizebox{0.97\textwidth}{!}{%
            \begin{tikzpicture}[every node/.style={inner sep=0,outer sep=0}]%
                \node[anchor = north east] at(0, 0){%
                    \includegraphics[scale=0.5, trim=5mm 5mm 5mm 5mm]{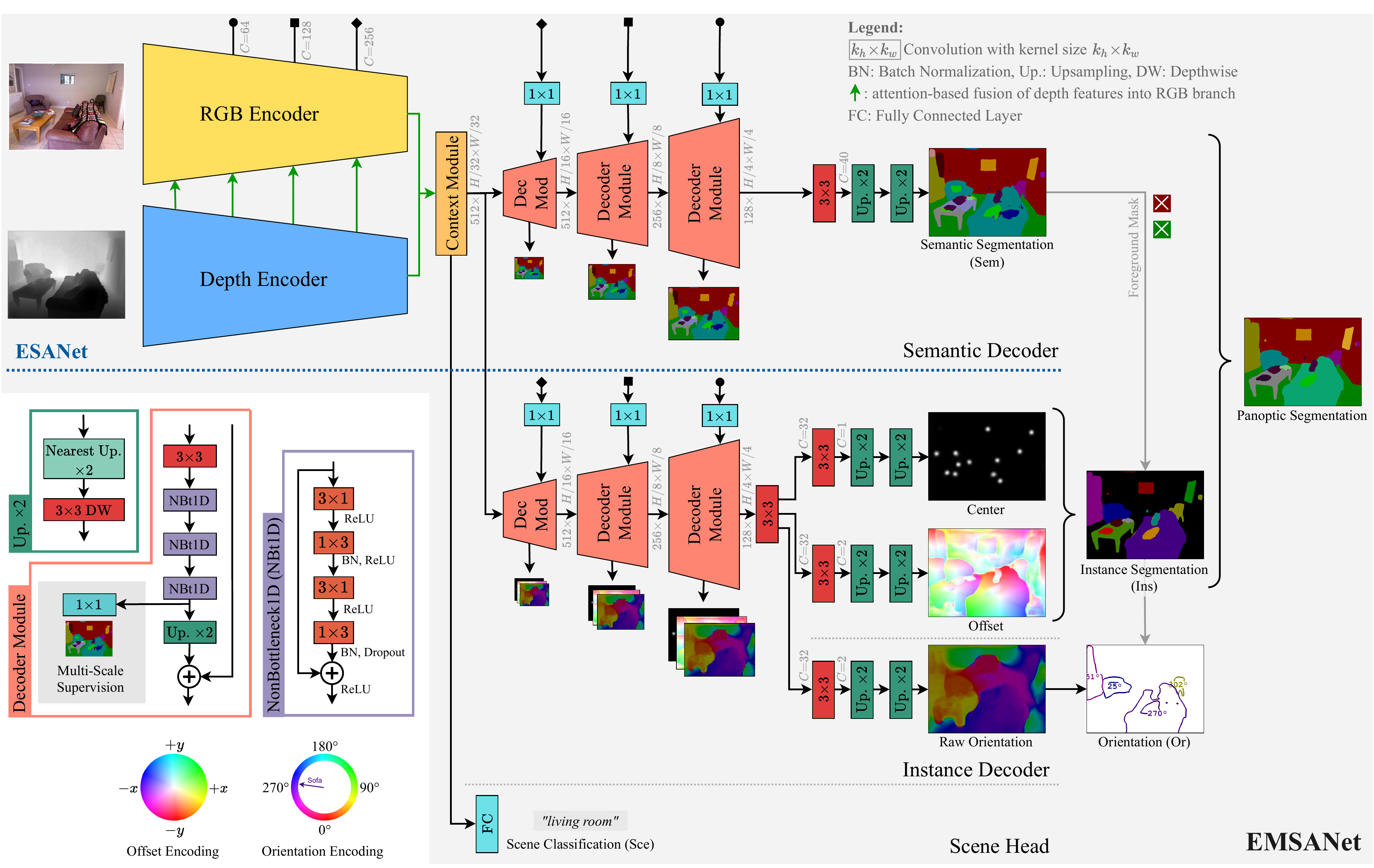}%
                };%
            \end{tikzpicture}%
        }%
    \vspace{-1mm}%
    \caption{%
        Architecture of our Efficient Multi-task Scene Analysis Network (EMSANet) extending ESANet~\cite{esanet2021icra} for semantic segmentation (top) with an additional decoder for instance segmentation and instance orientation estimation as well as a head for scene classification. %
        See Fig.~\ref{fig:iou-radar-chart} for semantic label colors.%
    }%
    \label{fig:architecture}%
    \vspace{-4mm}%
\end{figure*}

Our Efficient Multi-task Scene Analysis Network~(EMSANet) extends the encoder-decoder-based ESANet~\cite{esanet2021icra} for efficient RGB-D semantic segmentation.
As shown in Fig.~\ref{fig:architecture}~(top), ESANet features two identical encoders, one for processing RGB images and one for depth images.
For efficiency reasons, both encoders are based on a ResNet34~\cite{ResNet-cvpr2016} backbone.
For even faster inference and improved accuracy, the $3{\times}3$ convolutions are spatially factorized, resulting in the NonBottleneck1D block~(NBt1D)~\cite{ERFNet-its2018}~(see Fig.~\ref{fig:architecture} violet).
At each resolution stage of the encoders, an attention-based mechanism is used to fuse depth features into the RGB branch enhancing its representation with additional geometric information.
After the last fusion, a context module similar to the Pyramid Pooling Module in PSPNet~\cite{PSPNet-cvpr2017} is attached.
It incorporates context information at multiple scales using several branches with different pooling sizes~(see~\cite{esanet2021icra} for details).
The decoder is comprised of three decoder modules~(see Fig.~\ref{fig:architecture} light red), each is refining and upsampling the intermediate feature maps to gradually restore input resolution.
This is done by a $3{\times}3$ convolution followed by three NonBottleneck1D blocks and a final learned upsampling by a factor of two.
The learned upsampling~(see Fig.~\ref{fig:architecture} dark green) is initialized to mimic bilinear upsampling first.
However, as its weights are not fixed, the network is able to learn to combine adjacent features in a more
useful manner during training.
Additional encoder-decoder skip connections further help to restore spatial details that were lost during downsampling in the encoders.
Following the last decoder module, a $3{\times}3$ convolution maps the features to semantic classes.
Finally, two additional learned upsamplings restore the input resolution.
The entire network is trained end-to-end with additional side outputs and multi-scale supervision as depicted in Fig.~\ref{fig:architecture}.

ESANet builds a strong and efficient baseline for semantic segmentation. 
However, the architecture is specifically tailored for semantic segmentation. 
In order to improve its generalization capability for the remaining tasks, we further add a slight dropout with rate of 0.1 to all NonBottleneck1D blocks. 
Furthermore, we change the initialization in all \mbox{RGB-D} fusion modules to He-initialization~\cite{HeInit-iccv2015} and force zero-initialization~\cite{ZeroInit-arxiv2017} in all NonBottleneck1D blocks.
Finally, to incorporate the loss of other tasks more effectively, we do not reduce the accumulated loss by the sum of the applied semantic class weights but only by the number of all pixels across all outputs of the network.

Next, we present the extension to a multi-task network.

\subsection{Panoptic Segmentation}
\label{sec:main:panoptic_segmentation}
For panoptic segmentation, a second decoder for instance segmentation is required.
As shown in Fig.~\ref{fig:architecture} (middle), our instance decoder follows the same architecture as the semantic decoder except the task-specific heads.
The instance encoding follows the implementation of Panoptic DeepLab~\cite{PanopticDeeplab-cvpr2020}:~Instances are represented by their center of mass encoded as small 2D Gaussian within a heatmap, similar to keypoint estimation in other domains~\cite{OpenPose-cvpr2017}.
With an additional head, the instance decoder also predicts offset vectors for each pixel pointing towards a corresponding instance center in x and y direction.
As instances are only required for pixels belonging to thing classes -- i.e., all classes except wall, floor, and ceiling -- a corresponding foreground mask is derived from the semantic segmentation.
All thing pixels are then grouped into class-agnostic instances by combining both instance centers and offset predictions.
The semantic class for each instance is derived by a majority vote from the semantic segmentation.

Similar to Panoptic DeepLab, we use MSE loss for center prediction and L1 loss for offset prediction.
However, unlike Panoptic DeepLab, we mask predicted centers using the ground-truth instance mask instead of the thing-class mask to account for missing instance annotations in the ground truth. 
We also adopt their postprocessing, including thresholding and keypoint non-maximum suppression using max-pooling for the centers and the final merging of instance and semantic segmentation putting more focus on instances.
However, we faced some problems when applying their training regime. 
Panoptic DeepLab uses linear outputs for estimating both centers and absolute offsets.
This results in losses being unbounded and quite imbalanced, and, thus, requires a carefully tuned initialization and loss weights such as $200{\,:\,}0.1$ for center${\,:\,}$offsets as used in their implementation.
Moreover, absolute offset vectors do not generalize to varying input resolutions.
To address these issues, we use sigmoid activation for instance centers and a tanh activation for encoding relative instance offsets.
Thus, the outputs are bounded within $[0, 1]$ and $[-1, 1]$, respectively.
We observed great improvements in terms of stability during optimization and performance for instance segmentation.

\subsection{Instance Orientation Estimation}
\label{sec:main:orientation_estimation}
Our approach further predicts the orientation for instances of thing classes relevant for our indoor scenario, i.e., cabinet, bed, chair, sofa, bookshelf, shelves, dresser, refrigerator, tv, person, nightstand, and toilet.
The orientation is crucial when approaching objects from the right direction (e.g., chairs or persons) or to restrict waiting positions (e.g., the robot should not wait in the
sightline to a TV or in front of a freestanding chair or cabinet).
To accomplish this, as shown in Fig.~\ref{fig:architecture}~(middle), our instance decoder also predicts orientations as continuous angles around the axis perpendicular to the ground (see bottom-left legend for orientation encoding in Fig.~\ref{fig:architecture}).
Instead of relying on a patch-based orientation estimation, we follow our dense prediction design and propose to predict the orientation for all pixels of an instance.
This way, the instance awareness of our instance decoder can further be strengthened.
Moreover, to determine an instance's orientation, we are able to average multiple predictions approximating an ensemble effect.
We use the biternion encoding~\cite{Biternion-gcpr2015} and the von Mises loss function~\cite{Biternion-gcpr2015} to account for the periodicity of angles and to avoid discontinuities in the loss.

\subsection{Scene Classification}
\label{sec:main:scene_classification}
For scene classification, as shown in Fig.~\ref{fig:architecture}~(bottom), we simply apply a fully-connected layer on top of the context \mbox{module}.
However, as scene classification requires global context, we connect the fully-connected layer directly to the global average pooled branch of the context module.
Due to the noisy nature of scene classes, we further utilize label smoothing during training.
\section{Datasets}
\label{sec:datasets}
Training our proposed multi-task approach is challenging as it requires comprehensive data providing annotations for all tasks.
Moreover, our approach relies on both RGB and depth images as input.
Based on these requirements and our application scenario, below, we examine common RGB-D datasets for their suitability.
Furthermore, we describe how we enriched these datasets to enable training our multi-task approach.
Additional annotations are publicly available.

\textit{NYUv2:} \enspace %
The NYUv2 dataset~\cite{NYUv2-eccv2012} provides dense annotations for both semantic and instance segmentation.
For semantics, we use the common 40 classes setting.
However, this may lead to very small instances getting assigned to misleading classes, e.g. door knobs are assigned to cabinet, dresser, or nightstand. 
To avoid such bad assignments, we restrict instances to have at least an area of 
0.25\% of the image area.
For enabling panoptic segmentation, we declare wall, floor, and ceiling to be background~(stuff classes) and consider the remaining classes as thing classes.
In addition to these dense annotations, NYUv2 also provides ground-truth labels for scene classification.
However, annotations for instance orientations are missing so far. %
Therefore, we manually annotated the orientation for instances of the semantic classes mentioned in Sec.~\ref{sec:main:orientation_estimation}.
Due to perspective distortions, exact annotation of the orientation as egocentric angle around the axis perpendicular to the ground is not possible in the pure RGB image, which is why we annotated them directly in a point cloud.

\textit{SUNRGB-D:} \enspace %
The SUNRGB-D dataset~\cite{SUNRGBD-cvpr2015} combines multiple indoor \mbox{RGB-D} datasets, including NYUv2, and enriches them with additional annotations, making SUNRGB-D to be one of the most important datasets for real-world applications.
The dataset comes with annotations for scene classification and semantic segmentation.
Compared to NYUv2, the last three semantic filling classes, i.e., otherstructure, otherfurniture, and otherprop, are omitted and assigned to void.
Moreover, some semantic annotations in the NYUv2 part have further been assigned to the void class, resulting in minor differences to the original NYUv2 dataset.
Unfortunately, annotations for instance segmentation and instance orientation estimation are missing.
However, fortunately, SUNRGB-D also provides 3D bounding boxes, each with a class label and orientation, that can be used for instance extraction.
To obtain instances, we first created a mapping between semantic and box classes. 
Subsequently, we matched the box clusters with the semantic point cloud in 3D.
During matching, a unique instance label was assigned to all pixels belonging to the same semantic class.
This way, an instance mask as well as the orientation could be extracted for each bounding box.
However, one limitation of this approach is that not all objects within a scene were annotated with a 3D bounding box, making the instance masks more sparse.
To compensate this to some extent, we also merged the instance masks and orientations of NYUv2 back to \mbox{SUNRGB-D}.
For panoptic segmentation, we consider the same semantic classes to belong to stuff as for NYUv2.

\textit{Hypersim:} \enspace %
Unlike SUNRGB-D and NYUv2, Hypersim~\cite{hypersim-iccv2021} is a photo-realistic synthetic dataset.
For its creation, virtual cameras were placed in 461 professionally rendered 3D scenes, resulting in 77,400 samples, of which we use 72,419.
We blacklisted the remaining samples due to several scene or trajectory issues, i.e., void/single semantic label only, missing textures, or invalid depth.
Each sample provides an \mbox{RGB-D} image, a mask for semantics and instances, instance orientations, and a scene label.
However, the annotations for instance orientation are not consistent and, thus, cannot be used without further manual refinement.
As Hypersim adopts the NYUv2 semantic classes, the same partitioning for stuff and thing can be applied for panoptic segmentation.

\textit{Final remarks and further adjustments:} \enspace %
Due to the additional annotations, both NYUv2 and SUNRGB-D are suitable for training our full multi-task approach.
HyperSim provides high-quality synthetic data and, thus, is well suited for pretraining.
Tab.~\ref{tab:dataset} summarizes important statistics for all datasets used for training and evaluating our multi-task approach.
\begin{table}[!t]
\caption{
    Overview about the datasets used for our multi-task approach.
}
\scriptsize
\label{tab:dataset}
\centering
\vspace{-2.5mm}
\begin{tabular}{@{}llrrr@{}}
         \toprule
         & \textbf{Split} & \textbf{\#$\,$Images} & \textbf{\#$\,$Instances}    & \textbf{\#$\,$Orientations} \\ \midrule
\textbf{NYUv2}    & train & 795      & 12,092    & 2,696         \\
         & test  & 654      & 9,874    & 2,069         \\ \midrule %
\textbf{SUNRGB-D}  & train & 5,285    & 18,171    & 13,076       \\
         & test  & 5,050     & 16,961     & 12,440        \\ \midrule \addlinespace[.5em]
\textbf{Hypersim} & train & 57,443   & 3,009,566 & -             \\
         & valid & 7,286    & 261,677 & -             \\
         & test  & 7,690     & 374,052  & -             \\ \bottomrule
\end{tabular}
\vspace{-3.5mm}
\end{table}
For scene classification, we further created an own spectrum of classes that unifies the classes in all datasets and accounts for similar classes. 
The resulting spectrum is tailored for indoor applications and is comprised of the classes listed in Tab.~\ref{tab:scene}.
Note that the void class is used for images with unclear assignment that may disturb the learning process. 
Furthermore, images that show indoor scenes but cannot be assigned to one of the mentioned classes are considered as other indoor.

\begin{table}[t!]
\caption{
    Distribution of scene classes for all datasets and their splits.
}
\vspace{-2.5mm}
\scriptsize
\resizebox{\linewidth}{!}{%
\begin{tabular}{@{}l@{}rr@{\hspace{7mm}}rr@{\hspace{7mm}}rrr@{}}
\toprule%
                           & \multicolumn{2}{@{\hspace{5mm}}l}{\textbf{NYUv2}}   & \multicolumn{2}{@{\hspace{1mm}}l@{}}{\textbf{SUNRGB-D}} & \multicolumn{3}{c}{\textbf{Hypersim}}         \\%
                           & \textbf{~~train}   & \textbf{test}   & \textbf{train}   & \textbf{test}   & \textbf{train}   & \textbf{valid}   & \textbf{test}  \\%
                           \midrule
\textbf{void}              & 29                 & 0               & 745              & 507             & 1,370            & 0                & 0              \\%
\textbf{bathroom}          & 63                 & 58              & 331              & 293             & 3,378            & 400              & 300            \\%
\textbf{bedroom}           & 192                & 191             & 558              & 526             & 4,386            & 499              & 400            \\%
\textbf{dining room}       & 66                 & 55              & 311              & 296             & 1,894            & 200              & 100            \\%
\textbf{discussion room}   & 5                  & 0               & 691              & 753             & 1,013            & 100              & 0              \\%
\textbf{hallway}           & 0                  & 0               & 222              & 151             & 400              & 100              & 0              \\%
\textbf{kitchen}           & 125                & 110             & 288              & 297             & 6,221            & 400              & 600            \\%
\textbf{living room}       & 114                & 107             & 274              & 250             & 18,287           & 3,017            & 2,855          \\%
\textbf{office}            & 88                 & 78              & 820              & 792             & 5,026            & 100              & 389            \\%
\textbf{other indoor}      & 113                & 55              & 1,041            & 1,180           & 15,068           & 2,370            & 2,846          \\%
\textbf{stairs}            & 0                  & 0               & 4                & 5               & 400              & 100              & 200            \\%
\bottomrule%
\end{tabular}%
}%
\label{tab:scene}
\vspace{-5mm}
\end{table}

\section{Experiments}
\label{sec:experiments}
We evaluate our approach in several settings on the indoor datasets NYUv2 and SUNRGB-D.
First, we use the smaller NYUv2 dataset to elaborate suitable hyperparameters and task weights. 
We establish single-task baselines for each task and, subsequently, compare them to several multi-task settings.
Finally, we extend our studies to SUNRGB-D and Hypersim to examine the applicability to larger datasets and the relevance of synthetic data for pretraining.

\subsection{Implementation Details}
\label{sec:experiments:implementation}
Our architecture as well as the pipelines for training and evaluation are implemented using PyTorch~\cite{pytorch-neurips2019}.
We used pretrained weights on ImageNet~\cite{ImageNet-ijcv2015} to initialize both encoders and trained each network for 500 epochs with a batch size of 8.
For optimization, we used SGD with momentum of 0.9 and a small weight decay of 0.0001.
For determining a suitable learning rate, we performed a grid search with values of \{0.00125, 0.0025, 0.005, 0.01, 0.02, 0.03, 0.04, 0.08, 0.12\}.
The learning rate was further adapted during training using a one-cycle learning rate scheduler.
To increase the number of samples, we augmented images using random scaling, cropping, and flipping. 
For RGB images, we further applied slight color jittering in HSV space.

For postprocessing instance centers, we first apply a threshold of 0.1 and max-pooling with pooling size of 17 to perform keypoint non-maximum suppression, and finally filter the top-64 instances.
The pooling size results in the network not being able to predict instance centers closer than 8 pixels away from each other.
However, for both NYUv2 and SUNRGB-D, this decision affects less than 1\% of the instances.
For further details and other hyperparameters, we refer to our implementation available on GitHub.

\begin{figure*}[t!]
	\centering%
	\begin{tikzpicture}[scale=0.99]%
		\node[anchor=north] at (-0.1, 0.3){\small{(a) Semantic Segmentation (Sem)}};%
	    \node[anchor=north] at (-0.4, 0){%
	        \includegraphics[width=8.8cm]{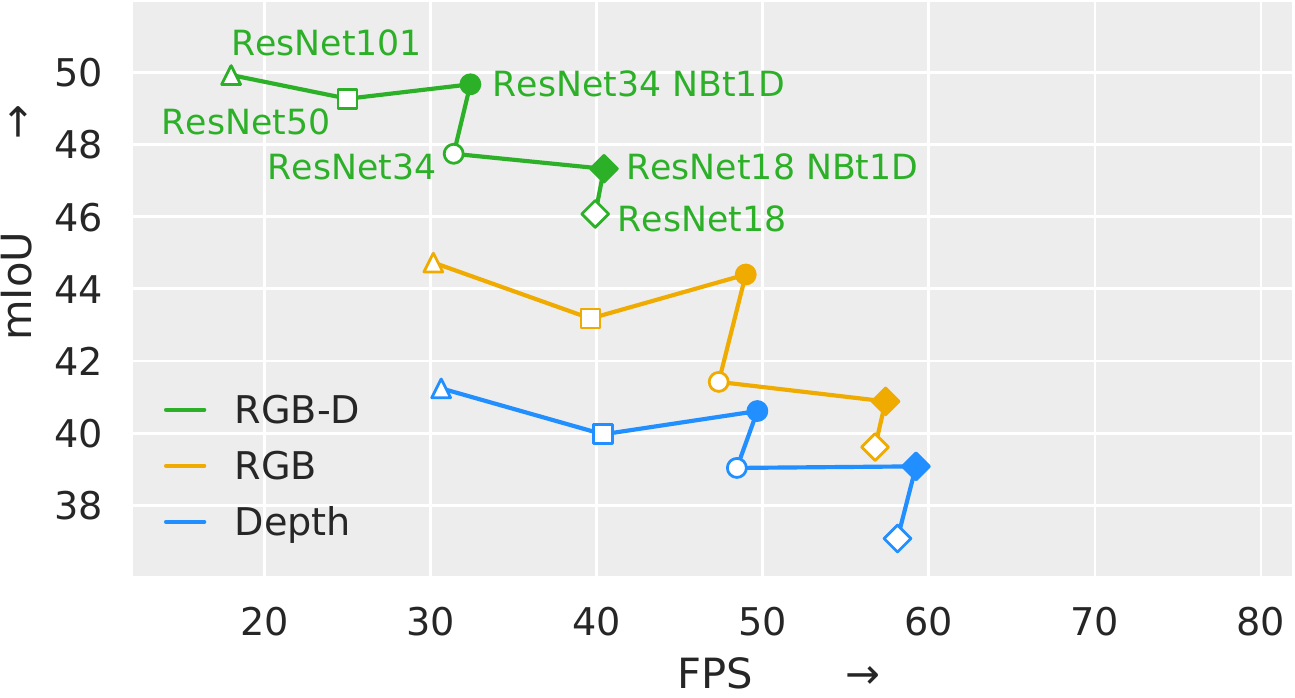}%
	    };
	    \node[anchor=north] at (9.2, 0.3){\small{(b) Instance Segmentation (Ins)}};%
        \node[anchor=north] at (8.8, 0){%
	        \includegraphics[width=8.8cm]{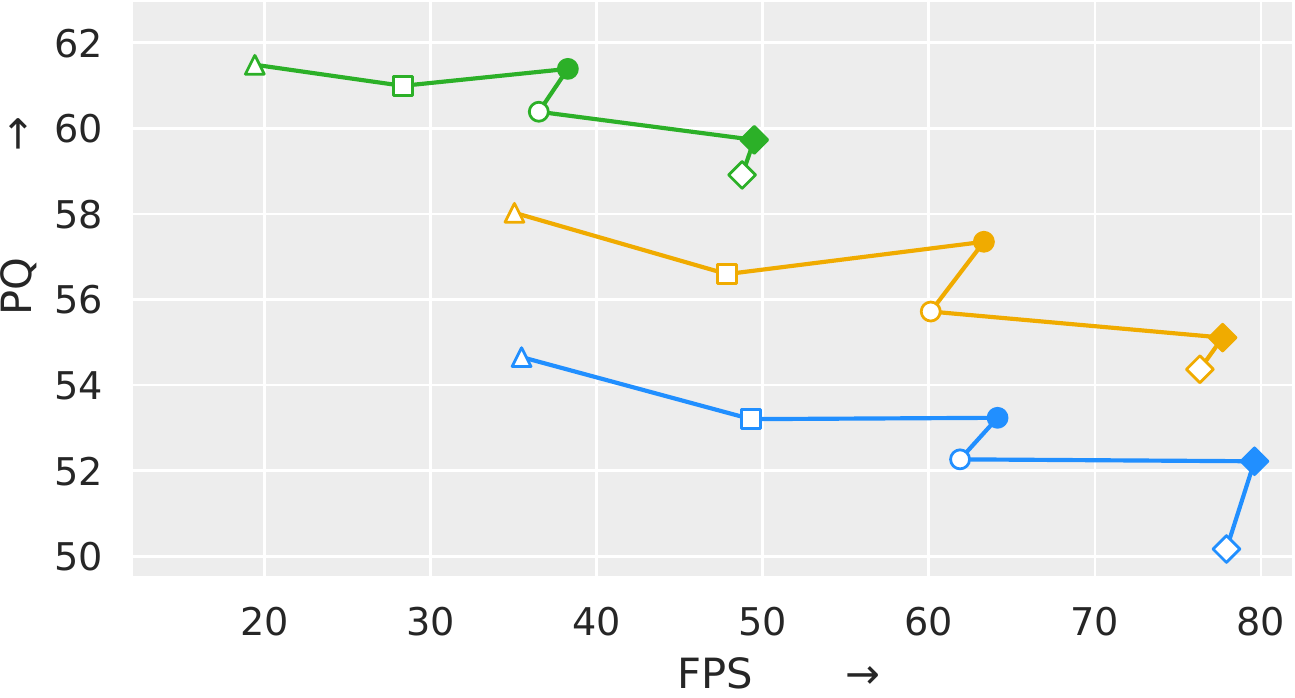}%
	    };
	    \node[anchor=north] at (-0.1, -4.9){\small{(c) Instance Orientation Estimation (Or)}};%
        \node[anchor=north] at (-0.4, -5.2){%
	        \includegraphics[width=8.8cm]{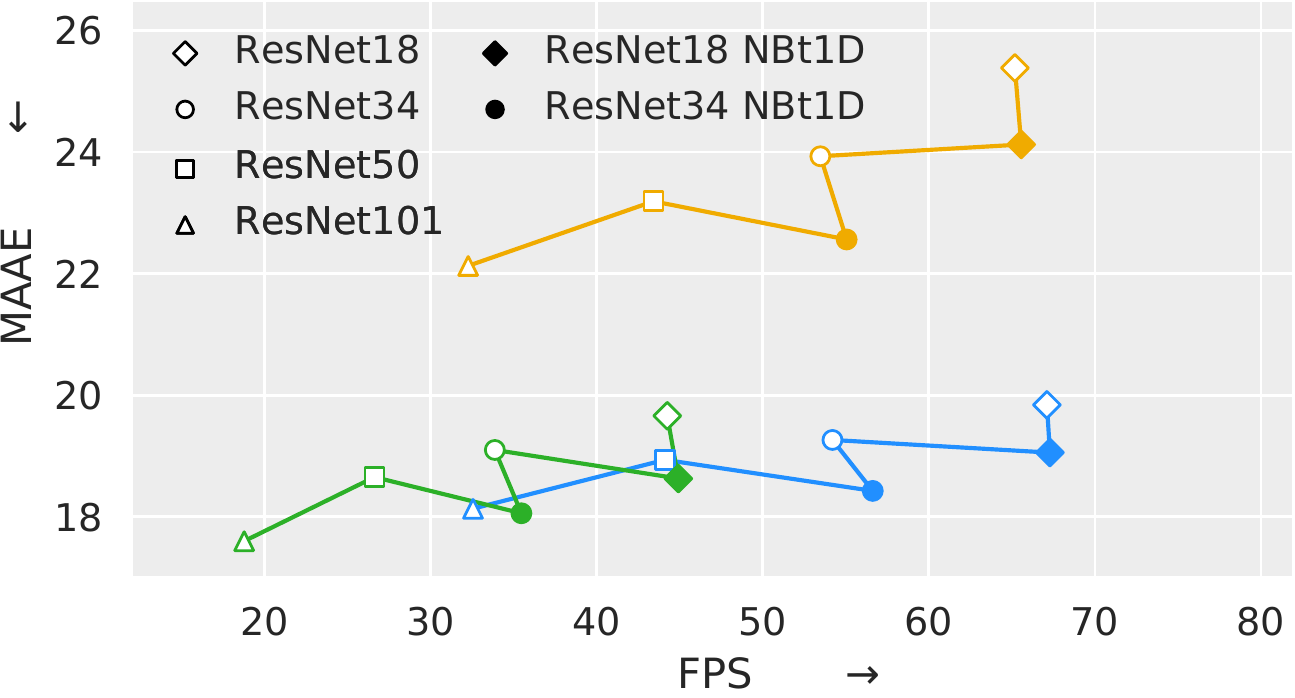}%
	    };
	    \node[anchor=north] at (9.2, -4.9){\small{(d) Scene Classification (Sce)}};%
        \node[anchor=north] at (8.8, -5.2){%
	        \includegraphics[width=8.8cm]{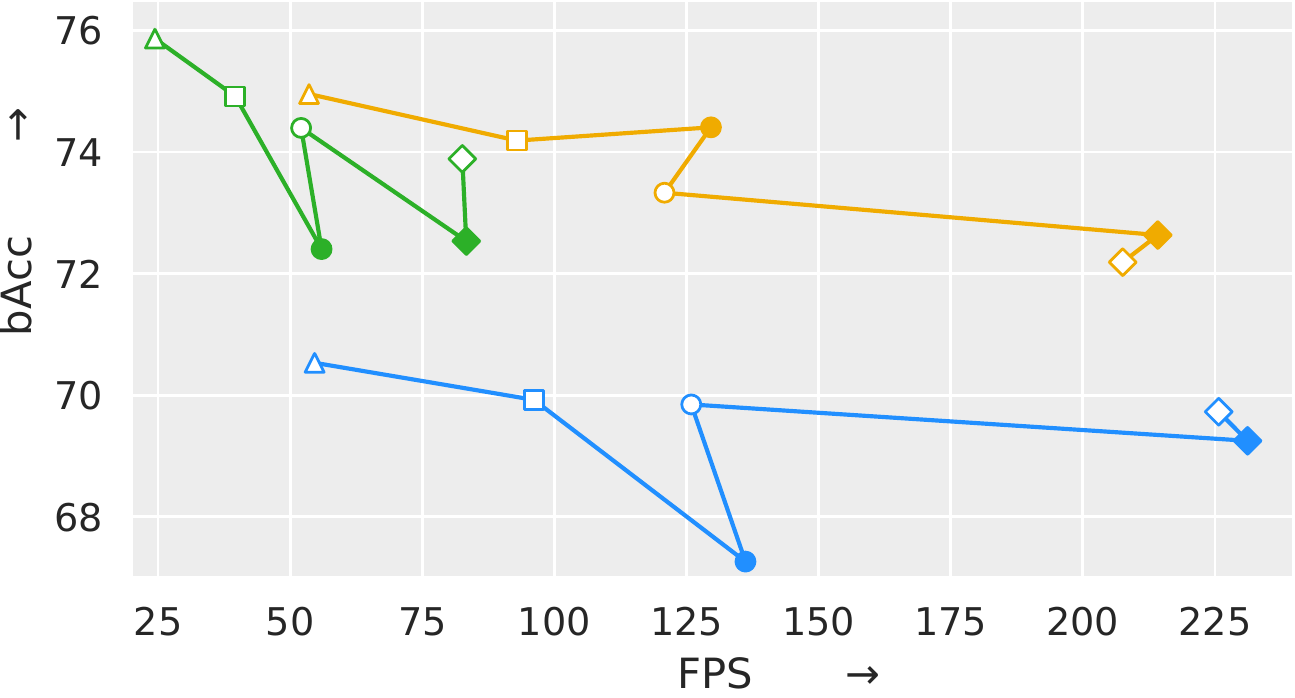}%
	    };
	\end{tikzpicture}%
	\vspace{-4mm}
	\caption{Results on NYUv2 test split when performing each task in a single-task setting with various backbones over the inference throughput in frames per second measured on an NIVIDA Jetson AGX Xavier~(Jetpack 4.6, TensorRT 8, Float16). See Sec.~\ref{sec:experiments:metrics} for metric abbreviations.}
	\label{fig:single_task}
	\vspace{-4mm}
\end{figure*}
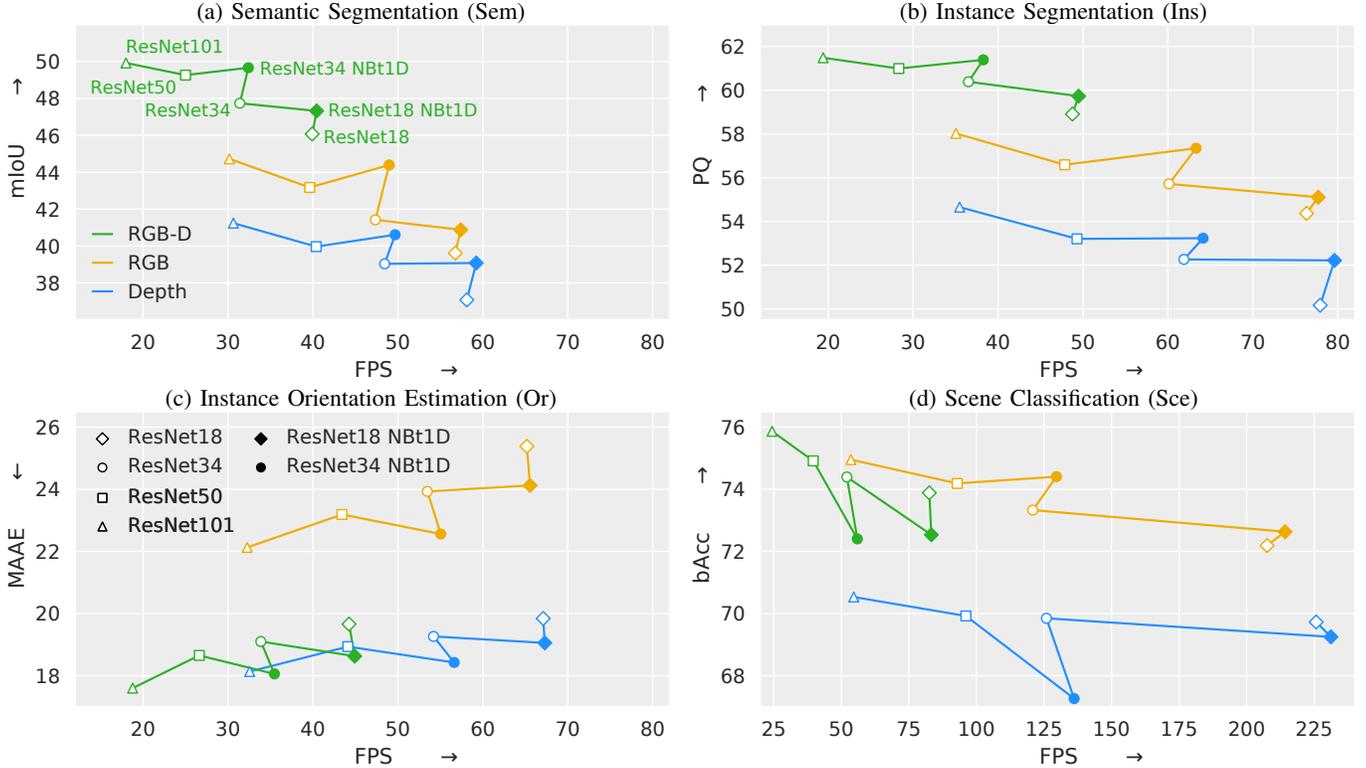

\subsection{Metrics}
\label{sec:experiments:metrics}
As we focus on fast inference, we do not apply any testing tricks such as horizontal flipping or multiscale inputs.
Before obtaining any performance metric for dense predictions, we resize the predictions to the full resolution of the ground truth.

\textit{Semantic Segmentation (Sem):} \enspace %
As common for this task, we use the mean intersection over union (mIoU).

\textit{Panoptic Segmentation:} \enspace %
The common metric for panoptic segmentation is the panoptic quality~(PQ)~\cite{PanopticSegmentation-cvpr2019}.
The panoptic quality PQ$_c$ for each class~$c$ is determined by the product of the recognition quality (RQ$_{c}$), i.e. the percentage of correctly detected instances similar to F$_1$-score, and the segmentation quality~(SQ$_c$), i.e., the segmentation accuracy similar to mIoU but only for matched segments. 
These metrics are typically averaged across all stuff~(st) and thing~(th) classes, resulting in RQ$^\text{st}$, SQ$^\text{st}$, PQ$^\text{st}$ and RQ$^\text{th}$, SQ$^\text{th}$, PQ$^\text{th}$, respectively.
It is also common to average the three metrics independently of stuff and things across all classes, resulting in RQ, SQ, and PQ.
Note that determining the metrics this way results in PQ to be typically not equal to RQ$\,\cdot\,$SQ.
For SUNRGB-D, we further ignore the classes floor mat and shower curtain as no instances of these classes occur in the test split.

\textit{Instance Segmentation (Ins):} \enspace %
For instance segmentation, we also stick to panoptic quality instead of reporting the average precision~(AP) as this would require assigning a confidence score to each instance~\cite{Coco-eccv2014}.
Moreover, AP and PQ track closely, which is why the latter also evaluates instance segmentation in a meaningful way~\cite{PanopticSegmentation-cvpr2019}.

\textit{Orientation Estimation (Or):} \enspace %
For evaluating instance orientations, we use the mean absolute angular error~(MAAE) in degrees similar to~\cite{DeepOrientation-iros2019, MTPersonPerception-iros2020}, i.e., in contrast to the other metrics, lower is better, and the maximum error is 180.
We report the MAAE for two settings: 1) independently of other tasks, i.e., using the ground-truth instances, and 2) for matched instances after panoptic merging.
Note that we do not penalize unmatched instances for the latter setting.

\textit{Scene Classification (Sce):} \enspace %
As the scene class labels are imbalanced, we evaluate scene classification with the balanced accuracy (bAcc).

\subsection{Single-task Setting}
\label{sec:experiments:single_task}
We aim at solving multiple tasks at once using a single neural network. 
To be able to elaborate whether this diminishes or boosts the performance of individual tasks, we first conducted experiments in a single-task setting.
Furthermore, the goal was to examine how the new tasks behave for different modalities, and how the additional network parts affect inference time on a mobile platform. 
Note that performing instance segmentation solely requires semantics and a foreground mask. 
We use the ground-truth semantic segmentation in this case, leading to RQ$^\text{st}$, SQ$^\text{st}$, and PQ$^\text{st}$ to always be equal to 1.
Moreover, as instance orientation estimation requires instance masks, we rely on the ground-truth instances as well.

Fig.~\ref{fig:single_task} shows the results of this set of experiments.
It becomes obvious that all tasks are able to benefit from incorporating complementary information processed in an additional depth branch instead of processing RGB only.
For semantic and instance segmentation, the results coincide with our findings in~\cite{esanet2021icra} that, whenever possible, processing both RGB and depth should be preferred over applying a more sophisticated single-modality RGB encoder.
The results further show that depth is crucial for estimating the orientation more accurately, and that RGB is essential for scene classification.
Finally, when comparing different backbones, it becomes obvious that, for all tasks except of scene classification, backbones featuring the NonBottleneck1D block~(NBt1D) lead to better results in terms of both performance and inference throughput compared to their counterparts with BasicBlock. 
Even more, ResNet34~NBt1D often competes with the more complex ResNet50 while enabling faster inference.
Therefore, we stick to ResNet34~NBt1D for the remaining experiments.

\subsection{Multi-task Setting}
\label{sec:experiments:multi_task}
Learning multiple tasks using a single neural network is challenging, as the tasks may influence each other.
Thus, tuning the task weights, which balance the losses to each other, is crucial. 
With uncertainty weighting~\cite{UncertaintyWeighting-cvpr2018}, GradNorm~\cite{GradNorm-icml2018}, Dynamic Weight Average~\cite{DynamicWeightAverage-cvpr2019}, and VarNorm~\cite{VarNorm-ral2021}, several approaches for determining the weights automatically have been proposed.
Unfortunately, none of them led to good performance in our scenario.
Therefore, we performed extensive experiments to determine suitable task weights.
We first combined two tasks at a time to elaborate essential relations between tasks. 
With these findings at hand, we were able to restrict the search space for the full multi-task setting.
Tab.~\ref{tab:results_nyuv2} summarizes the best results and compares them to their single-task counterparts of Fig.~\ref{fig:single_task}.
Furthermore, it lists the applied task weights, the learning rate, and the achieved frames per second on an NVIDIA Jetson AGX Xavier.

\begin{table*}[t!]
\center
\caption{%
    Results obtained on NYUv2 test split when training EMSANet with ResNet34~NBt1D backbone in various multi-task settings. %
    See Sec.~\ref{sec:experiments:metrics} for details on the reported metrics. %
    Panoptic results are obtained after merging semantic and instance prediction.
    Legend: italic:~metric used for determining the best checkpoint, *:~best result within the same run, Lr:~learning rate, pre.:~additional pretraining on Hypersim, FPS:~frames per second on an NVIDIA Jetson AGX Xavier (Jetpack 4.6, TensorRT 8, Float16).
}
\vspace{-5mm}
\resizebox{\textwidth}{!}{%
\scriptsize%
\begin{tabular}{@{}c@{\hspace{2.5mm}}lc@{\hspace{5mm}}c@{\hspace{5mm}}c@{\hspace{7mm}}c@{\hspace{2mm}}c@{\hspace{7mm}}c@{\hspace{7mm}}c@{\hspace{2mm}}c@{\hspace{2mm}}c@{\hspace{2mm}}c@{\hspace{1mm}}c@{\hspace{7mm}}c@{}}
\toprule
& & \multicolumn{1}{l}{} &  & \textbf{\begin{tabular}[c]{@{}c@{}}\color[HTML]{737373}Semantic\\ \color[HTML]{737373}Decoder\end{tabular}} & \multicolumn{2}{c@{\hspace{9mm}}}{\textbf{\begin{tabular}[c]{@{}c@{}}\color[HTML]{737373}Instance\\ \color[HTML]{737373}Decoder\end{tabular}}} & \textbf{\begin{tabular}[c]{@{}c@{}}\color[HTML]{5c5c5c}Scene\\ \color[HTML]{5c5c5c}Head\end{tabular}} & \multicolumn{5}{c@{\hspace{9mm}}}{\textbf{\begin{tabular}[c]{@{}c@{}}\color[HTML]{737373}Panoptic Results\\ \color[HTML]{737373}(after merging)\end{tabular}}} &  \\
& & \textbf{Task Weights} & \textbf{Lr} & \textbf{mIoU}$\,{}^\uparrow$ & \textbf{PQ}$\,{}^\uparrow$ & \textbf{MAAE}$\,{}^\downarrow$ & \textbf{bAcc}$\,{}^\uparrow$ & \textbf{mIoU}$\,{}^\uparrow$ & \textbf{PQ}$\,{}^\uparrow$ & \textbf{RQ}$\,{}^\uparrow$ & \textbf{SQ}$\,{}^\uparrow$ & \textbf{MAEE}$\,{}^\downarrow$ & \textbf{FPS}$\,{}^\uparrow$ \\ \midrule
& \textbf{Semantic Segmentation (Sem)} & --- & 0.04 & %
    49.66 & --- & --- & --- & --- & --- & --- & --- & --- & 32.4 \\
& \textbf{Instance Segmentation (Ins)} & --- & 0.08 & %
    --- & 61.39 & --- & --- & --- & --- & --- & --- & --- & 38.3 \\
& \textbf{Orientation Estimation (Or)} & --- & 0.02 & %
    --- & --- & 18.06 & --- & --- & --- & --- & --- & --- & 35.5  \\
\color[HTML]{9B9B9B} \multirow{-4}{*}{\rotatebox{90}{\textbf{ST}}} & \textbf{Scene Classification (Sce)} & --- & 0.02 & %
    --- & --- & --- & 72.40 & --- & --- & --- & --- & --- & 55.9 \\ \midrule
& \textbf{Sem$\,$+$\,$Sce} & $3{\,:\,}1$ & 0.02 & %
    \emph{49.57} & --- & --- & 74.29 & --- & --- & --- & --- & --- & 32.2 \\
\color[HTML]{9B9B9B} \multirow{-2}{*}{\rotatebox{90}{\textbf{\begin{tabular}[c]{@{}c@{}}MT\\[-0.5mm] \uproman{1}\end{tabular}}}} &  {\color[HTML]{9B9B9B} \textbf{Sem$\,$+$\,$Sce$\,$*}} & {\color[HTML]{9B9B9B} } &  & %
    {\color[HTML]{9B9B9B} \emph{49.57}} & {\color[HTML]{9B9B9B} ---} & {\color[HTML]{9B9B9B} ---} & {\color[HTML]{9B9B9B} 77.30} & {\color[HTML]{9B9B9B} ---} & {\color[HTML]{9B9B9B} ---} & {\color[HTML]{9B9B9B} ---} & {\color[HTML]{9B9B9B} ---} & {\color[HTML]{9B9B9B} ---} &  \\%
\midrule
& \textbf{Sem$\,$+$\,$Ins} & $1{\,:\,}3$ & 0.03 & %
    50.22 & 60.90 & --- & --- & 50.24 & \emph{43.74} & 52.46 & 82.43 & --- & 25.8 \\
\color[HTML]{9B9B9B} \multirow{-2}{*}{\rotatebox{90}{\textbf{\begin{tabular}[c]{@{}c@{}}MT\\[-0.5mm] \uproman{2}\end{tabular}}}} & {\color[HTML]{9B9B9B} \textbf{Sem$\,$+$\,$Ins$\,$*}} & {\color[HTML]{9B9B9B} } &  & %
    {\color[HTML]{9B9B9B} 50.22} & {\color[HTML]{9B9B9B} 61.61} & {\color[HTML]{9B9B9B} ---} & {\color[HTML]{9B9B9B} ---} & {\color[HTML]{9B9B9B} 50.54} & {\color[HTML]{9B9B9B} \emph{43.74}} & {\color[HTML]{9B9B9B} 52.50} & {\color[HTML]{9B9B9B} 82.63} & {\color[HTML]{9B9B9B} ---} &  \\%
\midrule
& \textbf{Ins$\,$+$\,$Or} & $3{\,:\,}1$ & 0.04 & %
    --- & \emph{59.72} & 17.66 & --- & --- & --- & --- & --- & --- & 35.5 \\
\color[HTML]{9B9B9B} \multirow{-2}{*}{\rotatebox{90}{\textbf{\begin{tabular}[c]{@{}c@{}}MT\\[-0.5mm] \uproman{3}\end{tabular}}}} & {\color[HTML]{9B9B9B} \textbf{Ins$\,$+$\,$Or$\,$*}} & {\color[HTML]{9B9B9B} } &  & %
    {\color[HTML]{9B9B9B} ---} & {\color[HTML]{9B9B9B} \emph{59.72}} & {\color[HTML]{9B9B9B} 17.53} & {\color[HTML]{9B9B9B} ---} & {\color[HTML]{9B9B9B} ---} & {\color[HTML]{9B9B9B} ---} & {\color[HTML]{9B9B9B} ---} & {\color[HTML]{9B9B9B} ---} & {\color[HTML]{9B9B9B} ---} &  \\%
\midrule
& \textbf{Sem$\,$+$\,$Sce$\,$+$\,$Ins$\,$+$\,$Or} & $1{\,:\,}0.25{\,:\,}3{\,:\,}1$ & 0.04 & %
    50.97 & 61.35 & 19.01 & 76.46 & 50.54 & \emph{43.56} & 52.20 & 82.48 & 16.38 & 24.5 \\
\color[HTML]{9B9B9B} \multirow{-2}{*}{\rotatebox{90}{\textbf{\begin{tabular}[c]{@{}c@{}}MT\\[-0.5mm] \uproman{4}\end{tabular}}}} & {\color[HTML]{9B9B9B} \textbf{Sem$\,$+$\,$Sce$\,$+$\,$Ins$\,$+$\,$Or$\,$*}} & {\color[HTML]{9B9B9B} } &  & %
    {\color[HTML]{9B9B9B} 51.15} & {\color[HTML]{9B9B9B} 61.53} & {\color[HTML]{9B9B9B} 18.93} & {\color[HTML]{9B9B9B} 78.18} & {\color[HTML]{9B9B9B} 51.31} & {\color[HTML]{9B9B9B} \emph{43.56}} & {\color[HTML]{9B9B9B} 52.27} & {\color[HTML]{9B9B9B} 82.70} & {\color[HTML]{9B9B9B} 15.76} &  \\%
\midrule\midrule
& \textbf{Sem$\,$+$\,$Sce$\,$+$\,$Ins$\,$+$\,$Or (pre.)} & $1{\,:\,}0.25{\,:\,}3{\,:\,}1$ & 0.01 & %
    53.34 & 64.41 & 18.84 & 75.25 & 53.79 & \emph{47.38} & 55.95 & 83.74 & 15.91 & 24.5 \\
\color[HTML]{9B9B9B} \multirow{-2}{*}{\rotatebox{90}{\textbf{\begin{tabular}[c]{@{}c@{}}MT\\[-0.5mm] \uproman{5}\end{tabular}}}}  & {\color[HTML]{9B9B9B} \textbf{Sem$\,$+$\,$Sce$\,$+$\,$Ins$\,$+$\,$Or (pre.)$\,$*}} & \multicolumn{1}{l}{} &  & %
    {\color[HTML]{999999} 53.55} & {\color[HTML]{999999} 64.98} & {\color[HTML]{999999} 18.27} & {\color[HTML]{999999} 76.98} & {\color[HTML]{999999} 54.00} & {\color[HTML]{999999} \emph{47.38}} & {\color[HTML]{999999} 55.99} & {\color[HTML]{999999} 84.08} & {\color[HTML]{999999} 15.56} & \\%
\bottomrule
\end{tabular}
}
\label{tab:results_nyuv2}
\vspace{-4mm}
\end{table*}

\textit{Sem$\,$+$\,$Sce:} \enspace %
As shown in Tab.~\ref{tab:results_nyuv2}~(MT~\uproman{1}), combining both tasks requires a much larger weight for semantic segmentation to reach its single-task performance. 
However, even when putting more weight on semantic segmentation, scene classification benefits from such a setting, already closing the gap shown in Fig.~\ref{fig:single_task}~(d) for ResNet34~NBt1D and ResNet34.
This shows that knowledge about the individual parts of a scene is shared and helps to classify the scene.

\textit{Sem$\,$+$\,$Ins:} \enspace %
This setting allows obtaining panoptic results with predicted semantic segmentation for the first time.
As the semantic segmentation provides the semantics and the foreground mask for instance segmentation, combing these two tasks is crucial for our multi-task system.
As shown in Tab.~\ref{tab:results_nyuv2}~(MT~\uproman{2}), the best PQ is achieved by putting more focus on instance segmentation.
The mIoU obtained for the semantic decoder indicates that the network further benefits from performing both tasks in conjunction.
When keeping in mind, that the PQ for the instance decoder is computed using the ground-truth semantic segmentation, it is reasonable that the PQ for the panoptic results is lower.
With ground-truth semantic segmentation, the network reaches an RQ of $70.15$ and a SQ of $85.78$ (not listed in Tab.~\ref{tab:results_nyuv2}).
This shows that the drop in PQ is mainly due to a loss in RQ.
We observed that this is mostly caused by small instances, which are not part of the predicted foreground mask.

\textit{Ins$\,$+$\,$Or:} \enspace %
The results in Tab.~\ref{tab:results_nyuv2}~(MT~\uproman{3}) reveal that both tasks can be performed using a single decoder. 
Compared to the single-task baseline, this setting slightly improves orientation estimation, almost surpassing the reachable level for annotating orientations in 3D.
However, even when putting more weight on instance segmentation, we always observed a slight drop in PQ.

\begin{figure}[b!]
    \centering
    \vspace{-4mm}
    \includegraphics[width=\linewidth]{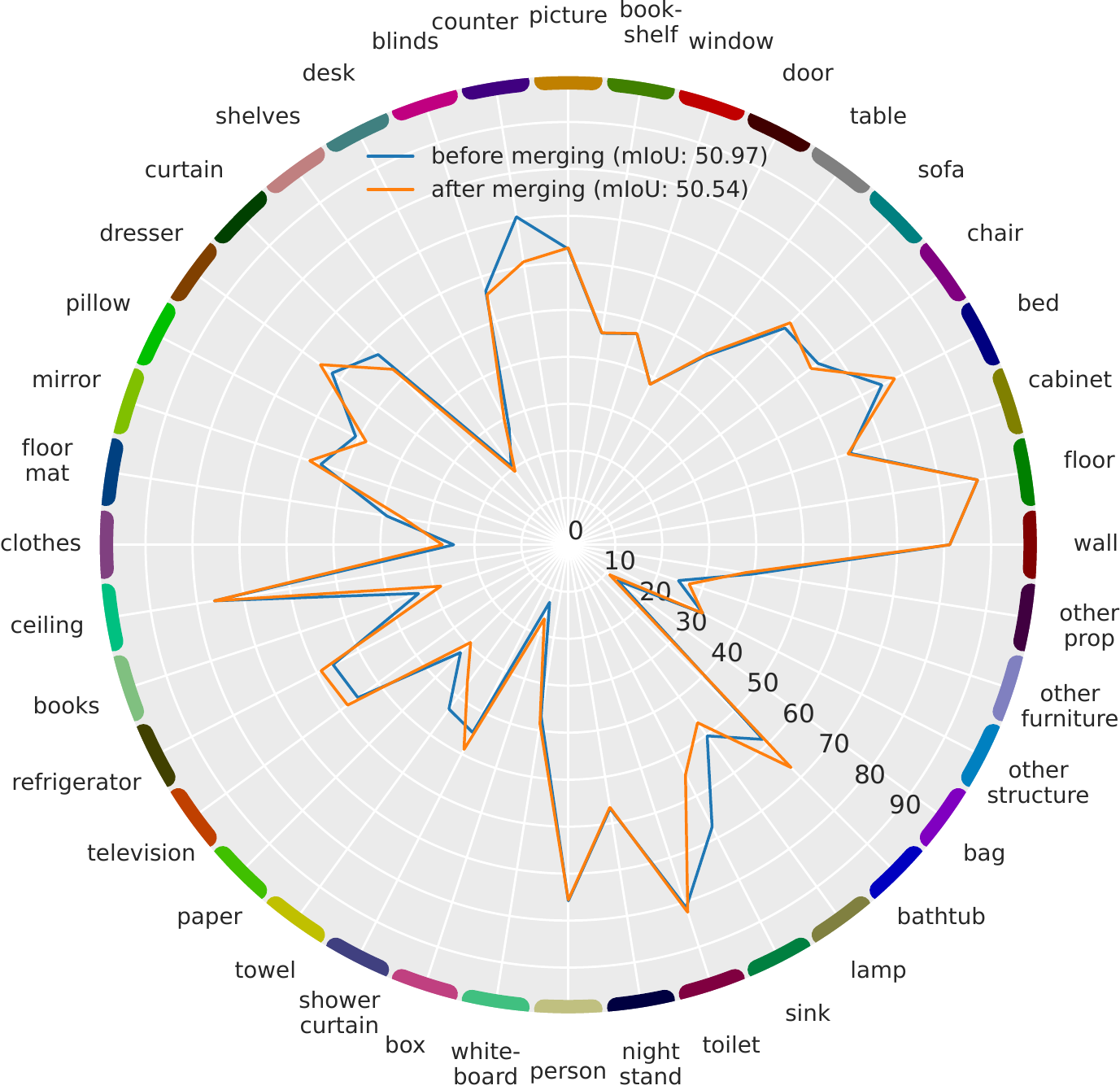}
    \vspace{-6mm}
    \caption{Semantic IoUs on NYUv2 test split for the full multi-task network (MT~\uproman{4}) before and after merging semantic and instance segmentation.}
    \label{fig:iou-radar-chart}
\end{figure}

\textit{Sem$\,$+$\,$Sce$\,$+$\,$Ins$\,$+$\,$Or:} \enspace %
With the findings of the aforementioned dual-task experiments at hand, we combined all tasks in a single neural network.
The best result is presented in Tab.~\ref{tab:results_nyuv2}~(MT~\uproman{4}).
It becomes obvious that both semantic segmentation and scene classification greatly benefit from the entire multi-task setting. 
Instance segmentation and instance orientation estimation almost reach the same level of accuracy as when performed in single-task settings.
The panoptic results,~i.e., after merging semantic and instance predictions, are similar to the multi-task setting MT~\uproman{2}. 
The mIoU obtained after merging is slightly lower than before merging but still at a similar level.
This indicates that the applied merging of both predictions with focus on instances does not diminish the semantic segmentation result.
The detailed breakdown of the IoUs in Fig.~\ref{fig:iou-radar-chart} shows that this holds true for almost all classes.
Finally, when taking a look at the results for orientation estimation, it can be seen that the orientation error after merging is lower than the error for the instance decoder.
However, this does not necessarily indicate better results, as the MAAE after panoptic merging only represents instances that could be matched.

\subsection{Results on SUNRGB-D}
\label{sec:experiments:results_sunrgbd}
\begin{figure}[b!]%
    \centering%
    \vspace{-5mm}%
    \resizebox{\linewidth}{!}{%
    \begin{tikzpicture}%
        \node[anchor=north] at (0, 0){%
	        \includegraphics[width=2.9cm]{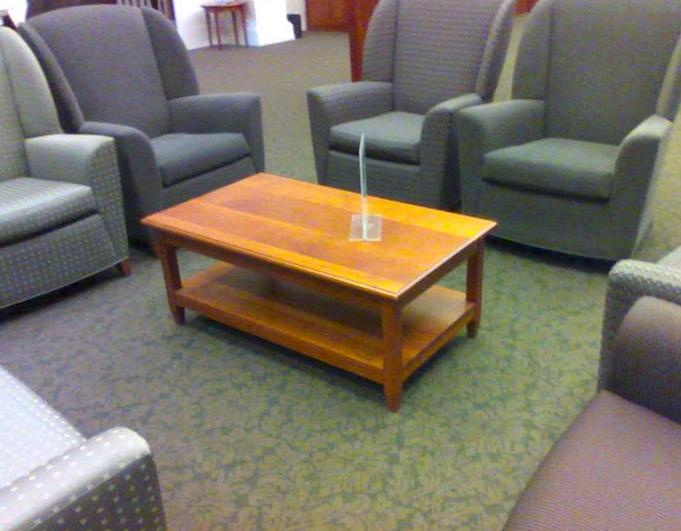}%
	    };%
	    \node[anchor=north] at (3, 0){%
	        \includegraphics[width=2.9cm]{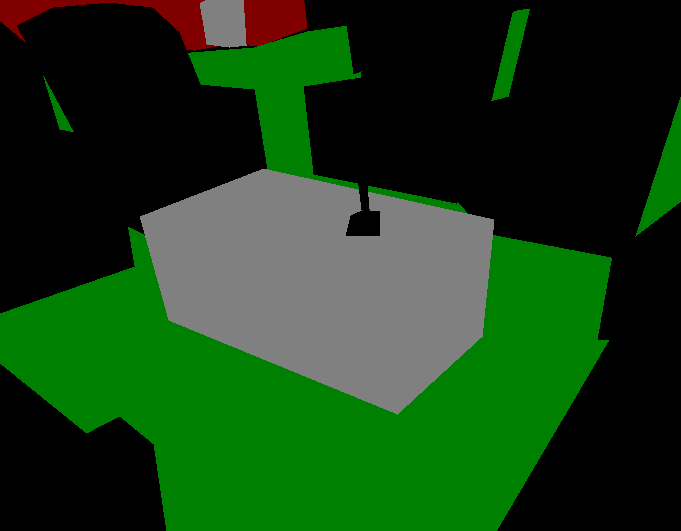}%
	    };%
	    \node[anchor=north] at (6, 0){%
	        \includegraphics[width=2.9cm]{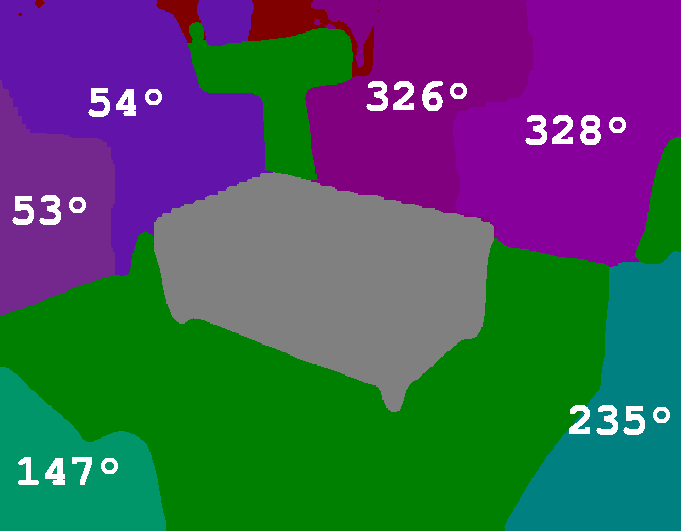}%
	    };%
	    \node[anchor=north] at (0, -2.4){%
	        \includegraphics[width=2.9cm]{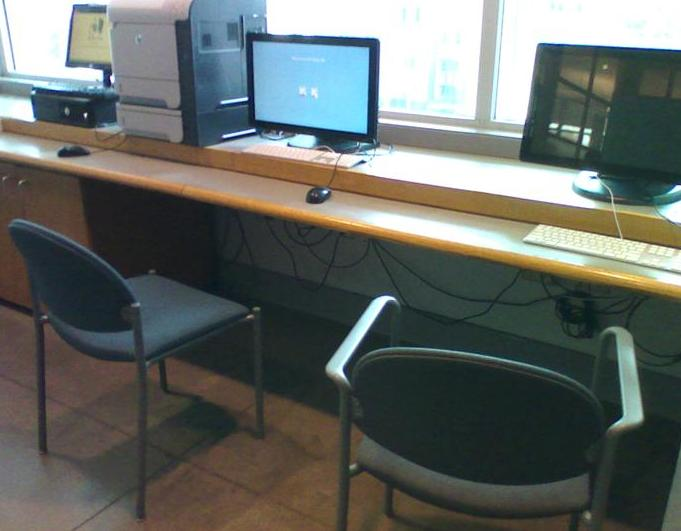}%
	    };%
	    \node[anchor=north] at (3, -2.4){%
	        \includegraphics[width=2.9cm]{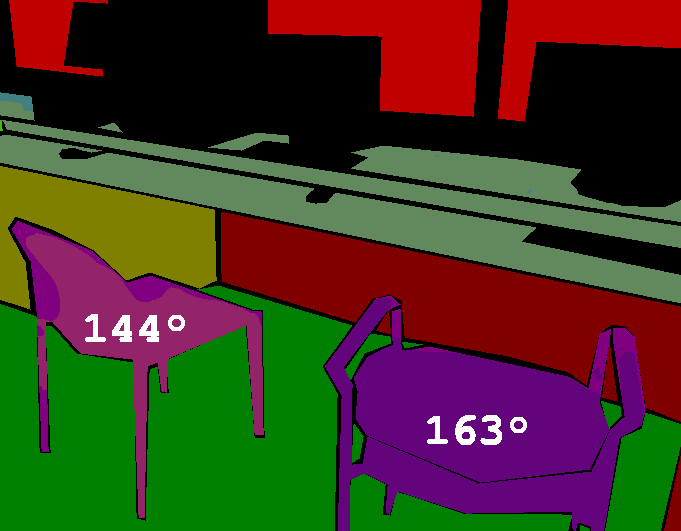}%
	    };%
	    \node[anchor=north] at (6, -2.4){%
	        \includegraphics[width=2.9cm]{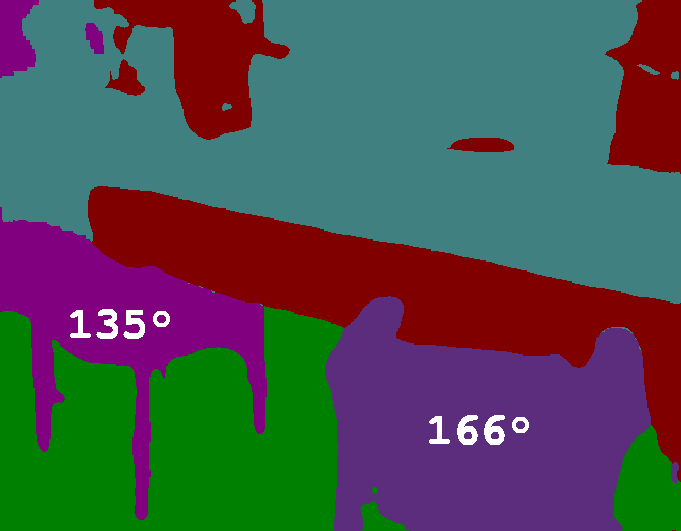}%
	    };%
    \end{tikzpicture}%
    }%
    \vspace{-4mm}%
    \caption{Qualitative results on SUNRGB-D test split highlighting faced challenges~(Sec~\ref{sec:experiments:results_sunrgbd}) as RGB image, ground-truth panoptic segmentation with orientations, and predicted panoptic segmentation with orientations.}
    \label{fig:sunrgbd_qualitative_results}
\end{figure}

After elaborating suitable multi-task parameters, we also applied our approach to the larger SUNRGB-D dataset. 
However, as instances in SUNRGB-D are more sparse, we put less weight on the instance decoder and used the mIoU for determining the best checkpoint.
The results are listed in Tab.~\ref{tab:results_sunrgbd}. 
Compared to the single-task baselines, our multi-task approach reaches slightly better performance for semantic segmentation and scene classification. 
The result for orientation estimation is still suitable for real-world application.
As shown in Fig.~\ref{fig:sunrgbd_qualitative_results}, the network generalizes well for instance segmentation even though annotations are much more sparse.
Note that the panoptic quality does not account for these areas as they are labeled as void class in the ground truth.
However, as shown in the last row, missing instance centers can still lead to assigning pixels far away to the same instance, lowering the PQ and mIoU after merging. 
We already observed a great benefit when masking centers based on the ground-truth instance mask during training, as proposed in Sec.~\ref{sec:main:panoptic_segmentation}. 
For real-world application, we further tackle this issue by thresholding predicted offsets after shifting and assign an unknown instance label if they are too far away from a center. 

\begin{table*}[!t]
\center
\caption{
    Results obtained on SUNRGB-D test split when training EMSANet with ResNet34~NBt1D backbone in both single-task and multi-task settings. %
    See Sec.~\ref{sec:experiments:metrics} for details on the reported metrics. %
    Panoptic results are obtained after merging semantic and instance prediction.
    Legend: italic:~metric used for determining the best checkpoint, *:~best result within the same run, Lr:~learning rate, pre.:~additional pretraining on Hypersim. %
}
\vspace{-2mm}
\scriptsize
\begin{tabular}{@{}c@{\hspace{2.5mm}}lc@{\hspace{5mm}}l@{\hspace{5mm}}c@{\hspace{7mm}}c@{\hspace{2mm}}c@{\hspace{7mm}}c@{\hspace{7mm}}c@{\hspace{2mm}}c@{\hspace{2mm}}c@{\hspace{2mm}}c@{\hspace{1mm}}c@{}}
\toprule
& & &  & \textbf{\begin{tabular}[c]{@{}c@{}}\color[HTML]{737373}Semantic\\ \color[HTML]{737373}Decoder\end{tabular}} & \multicolumn{2}{c@{\hspace{9mm}}}{\textbf{\begin{tabular}[c]{@{}c@{}}\color[HTML]{737373}Instance\\ \color[HTML]{737373}Decoder\end{tabular}}} & \textbf{\begin{tabular}[c]{@{}c@{}}\color[HTML]{5c5c5c}Scene\\ \color[HTML]{5c5c5c}Head\end{tabular}} & \multicolumn{5}{c@{\hspace{9mm}}}{\textbf{\begin{tabular}[c]{@{}c@{}}\hspace{6.5mm}\color[HTML]{737373}Panoptic Results\\ \hspace{6.5mm}\color[HTML]{737373}(after merging)\end{tabular}}}  \\
& & \textbf{Task Weights} & \textbf{~~Lr} & \textbf{mIoU}$\,{}^\uparrow$ & \textbf{PQ}$\,{}^\uparrow$ & \textbf{MAAE}$\,{}^\downarrow$ & \textbf{bAcc}$\,{}^\uparrow$ & \textbf{mIoU}$\,{}^\uparrow$ & \textbf{PQ}$\,{}^\uparrow$ & \textbf{RQ}$\,{}^\uparrow$ & \textbf{SQ}$\,{}^\uparrow$ & \textbf{MAEE}$\,{}^\downarrow$ \\ \midrule
& \textbf{Semantic Segmentation (Sem)} & --- & 0.005 & %
    48.23 & --- & --- & --- & --- & --- & --- & --- & --- \\
& \textbf{Instance Segmentation (Ins)} & --- & 0.01 & %
    --- & 60.99 & --- & --- & --- & --- & --- & --- & --- \\
& \textbf{Orientation Estimation (Or)} & --- & 0.005 & %
    --- & --- & 13.68 & --- & --- & --- & --- & --- & --- \\
\color[HTML]{9B9B9B} \multirow{-4}{*}{\rotatebox{90}{\textbf{ST}}} & \textbf{Scene Classification (Sce)} & --- & 0.001 & %
    --- & --- & --- & 58.66 & --- & --- & --- & --- & --- \\ \midrule
& \textbf{Sem$\,$+$\,$Sce$\,$+$\,$Ins$\,$+$\,$Or} & $1{\,:\,}0.25{\,:\,}2{\,:\,}0.5$ & 0.005 & %
    \emph{48.39} & 60.60 & 16.81 & 61.83 & 45.56 & 50.15 & 58.14 & 84.85 & 14.24 \\
\color[HTML]{9B9B9B} \multirow{-2}{*}{\rotatebox{90}{\textbf{\begin{tabular}[c]{@{}c@{}}MT\\[-0.5mm] \uproman{4}\end{tabular}}}}  & {\color[HTML]{9B9B9B} \textbf{Sem$\,$+$\,$Sce$\,$+$\,$Ins$\,$+$\,$Or$\,$*}} & {\color[HTML]{9B9B9B} } & \multicolumn{1}{l}{{\color[HTML]{9B9B9B} }} & %
    {\color[HTML]{9B9B9B} \emph{48.39}} & {\color[HTML]{9B9B9B} 61.48} & {\color[HTML]{9B9B9B} 16.70} & {\color[HTML]{9B9B9B} 62.66} & {\color[HTML]{9B9B9B} 45.66} & {\color[HTML]{9B9B9B} 50.53} & {\color[HTML]{9B9B9B} 58.66} & {\color[HTML]{9B9B9B} 85.20} & {\color[HTML]{9B9B9B} 14.15} \\ \midrule\midrule
& \textbf{Sem$\,$+$\,$Sce$\,$+$\,$Ins$\,$+$\,$Or (pre.)} & $1{\,:\,}0.25{\,:\,}2{\,:\,}0.5$ & 0.0025 & %
    \emph{48.47} & 64.24 & 18.40 & 57.22 & 44.18 & 52.84 & 60.67 & 86.01 & 14.10 \\
\color[HTML]{9B9B9B} \multirow{-2}{*}{\rotatebox{90}{\textbf{\begin{tabular}[c]{@{}c@{}}MT\\[-0.5mm] \uproman{5}\end{tabular}}}}  & {\color[HTML]{9B9B9B} \textbf{Sem$\,$+$\,$Sce$\,$+$\,$Ins$\,$+$\,$Or (pre.)$\,$*}}&  & \multicolumn{1}{l}{} & %
    {\color[HTML]{999999} \emph{48.47}} & {\color[HTML]{999999} 64.82} & {\color[HTML]{999999} 17.94} & {\color[HTML]{999999} 59.39} & {\color[HTML]{999999} 45.04} & {\color[HTML]{999999} 53.35} & {\color[HTML]{999999} 61.31} & {\color[HTML]{999999} 86.25} & {\color[HTML]{999999} 14.04} \\ \bottomrule
\end{tabular}%
\label{tab:results_sunrgbd}
\vspace{-4mm}
\end{table*}

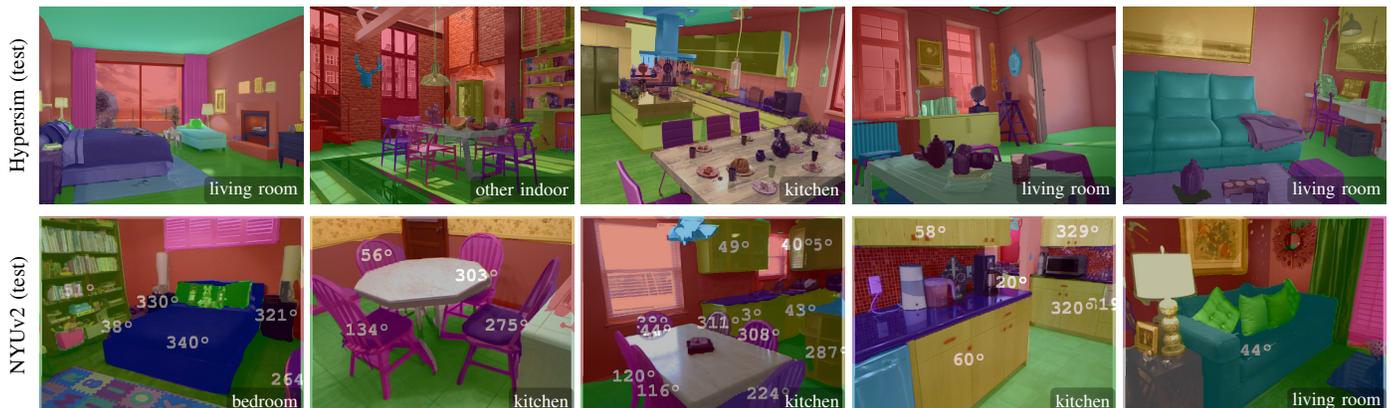
\begin{figure*}[b!]
    \centering
    \vspace{-5mm}
    \begin{tikzpicture}
        \tikzstyle{scene_label}=[anchor=north east, text=white, fill=black, fill opacity=0.3, text opacity=1, inner sep=2pt, rounded corners=2pt, minimum height=3.5mm, font=\scriptsize]
        \node[anchor=west, rotate=90] at (-0.15, 1.78){\footnotesize{Hypersim (test)}};%
        \node[anchor=west] at (0, 2.8){%
    	    \includegraphics[width=3.5cm]{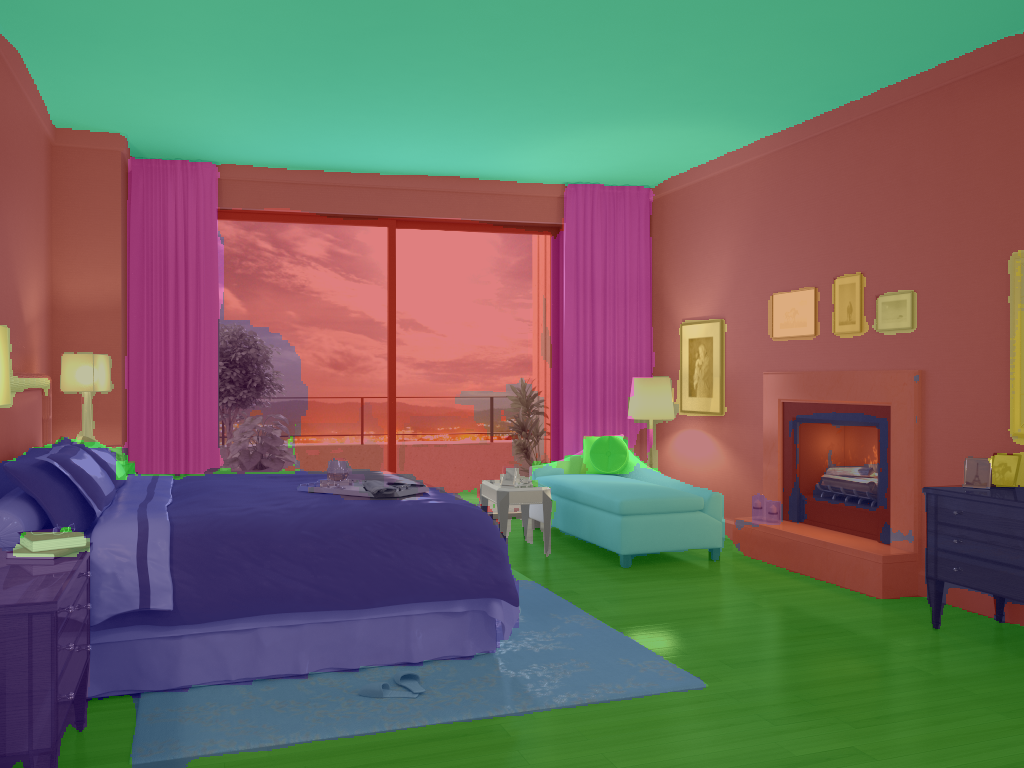}%
    	};%
    	\node[scene_label] at (3.63, 1.86){living room};%
    	\node[anchor=west] at (3.6, 2.8){%
    	    \includegraphics[width=3.5cm]{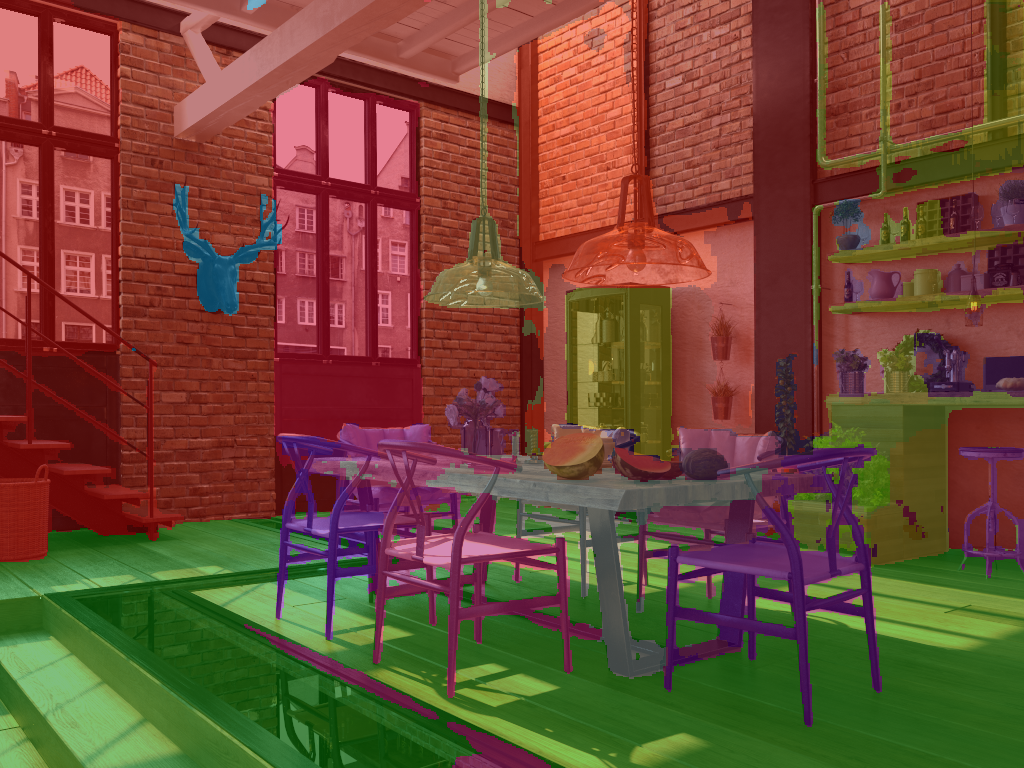}%
    	};%
    	\node[scene_label] at (7.23, 1.86){other indoor};%
    	\node[anchor=west] at (7.2, 2.8){%
    	    \includegraphics[width=3.5cm]{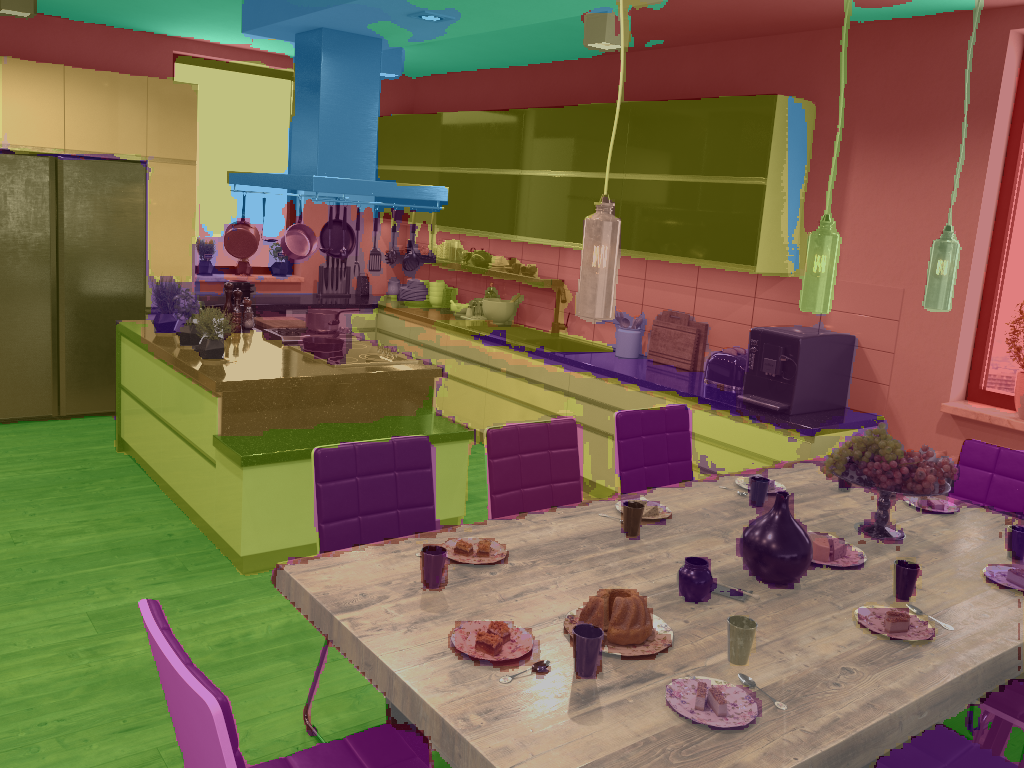}%
    	};%
    	\node[scene_label] at (10.83, 1.86){kitchen};%
    	\node[anchor=west] at (10.8, 2.8){%
    	    \includegraphics[width=3.5cm]{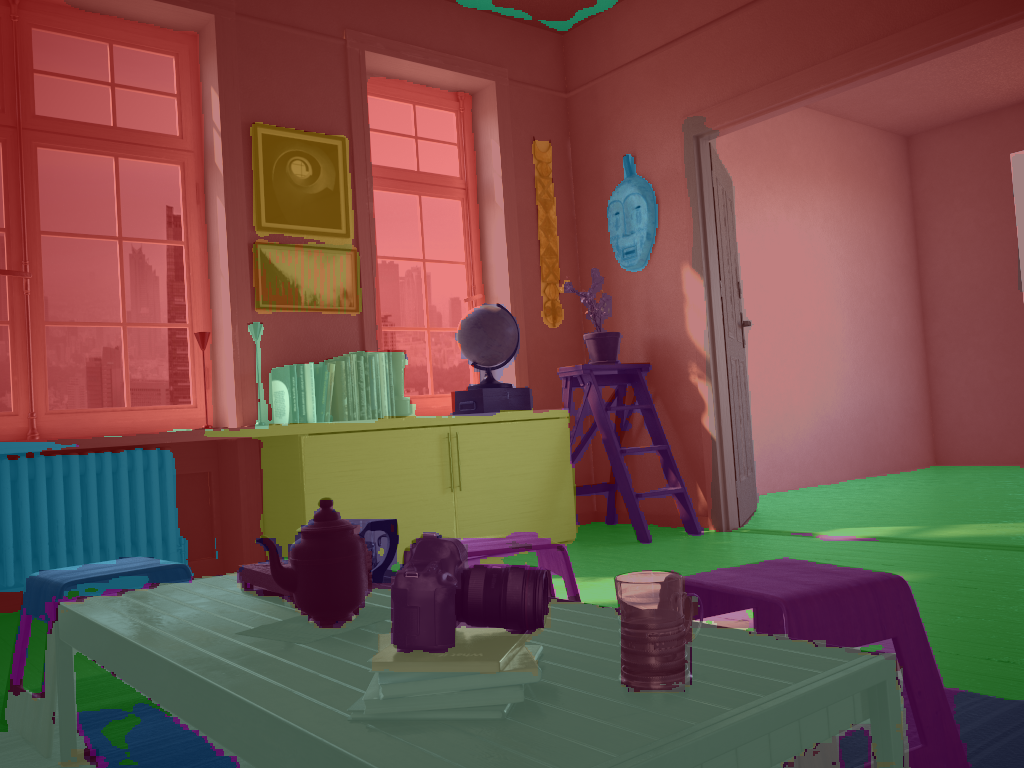}%
    	};%
    	\node[scene_label] at (14.43, 1.86){living room};%
    	\node[anchor=west] at (14.4, 2.8){%
    	    \includegraphics[width=3.5cm]{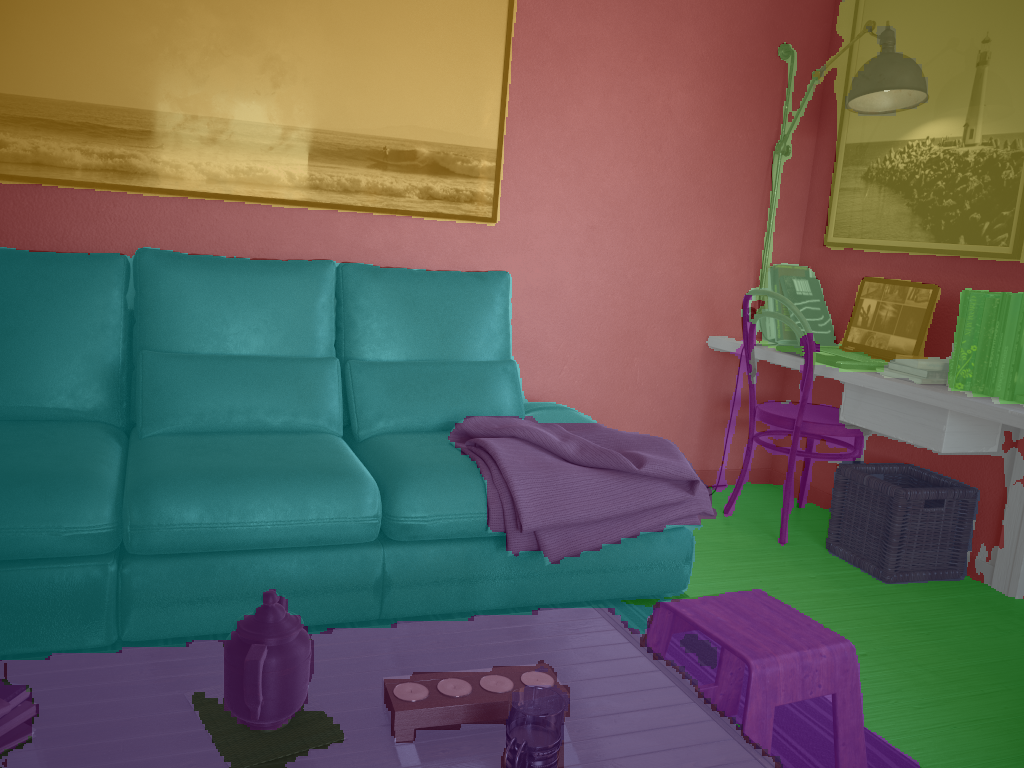}%
    	};%
    	\node[scene_label] at (18.03, 1.86){living room};%
        \node[anchor=west, rotate=90] at (-0.15, -0.9){\footnotesize{NYUv2 (test)}};%
        \node[anchor=west] at (0, 0){%
    	    \includegraphics[width=3.5cm]{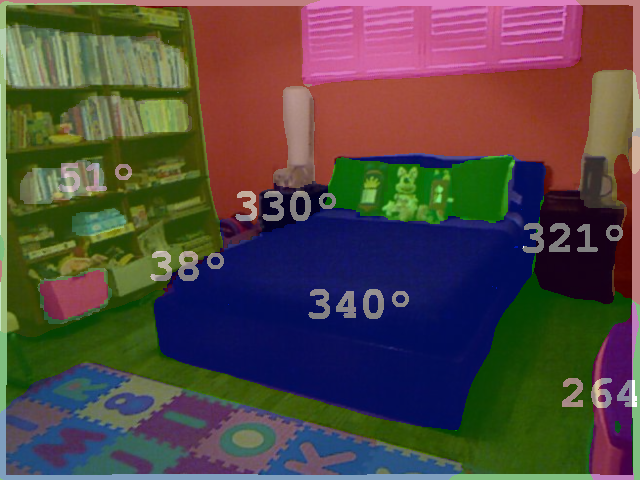}%
    	};%
    	\node[scene_label] at (3.63, -0.95){bedroom};%
    	\node[anchor=west] at (3.6, 0){%
    	    \includegraphics[width=3.5cm]{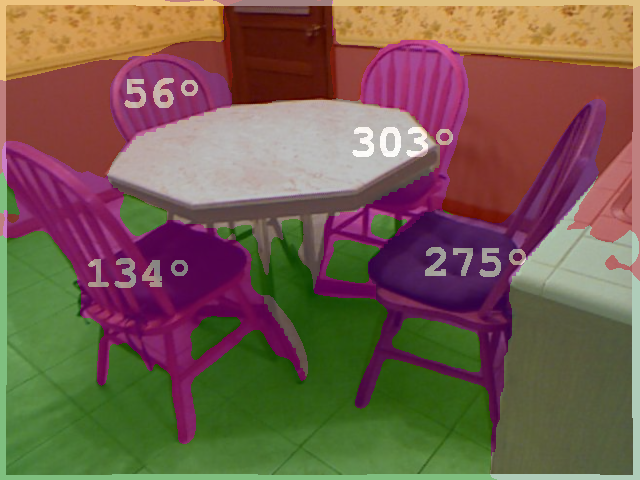}%
    	};%
    	\node[scene_label] at (7.23, -0.95){kitchen};%
    	\node[anchor=west] at (7.2, 0){%
    	    \includegraphics[width=3.5cm]{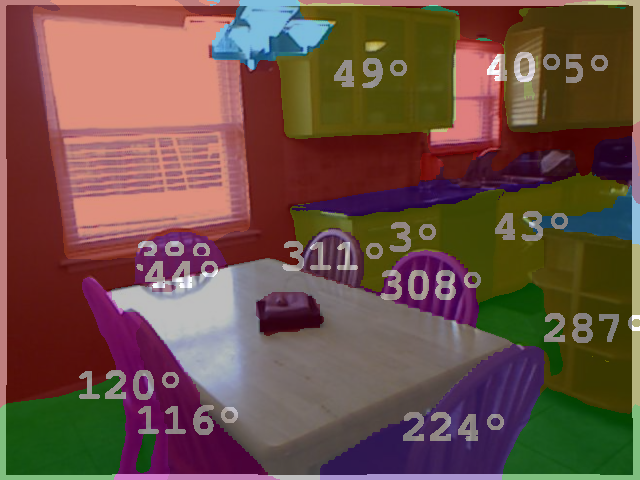}%
    	};%
    	\node[scene_label] at (10.83, -0.95){kitchen};%
    	\node[anchor=west] at (10.8, 0){%
    	    \includegraphics[width=3.5cm]{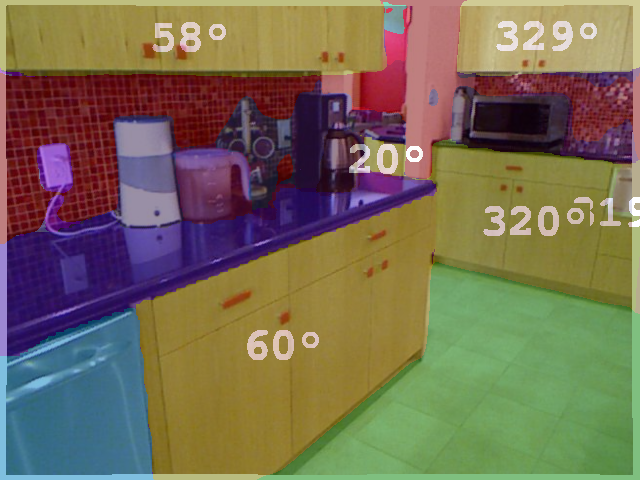}%
    	};%
    	\node[scene_label] at (14.43, -0.95){kitchen};%
    	\node[anchor=west] at (14.4, 0){%
    	    \includegraphics[width=3.5cm]{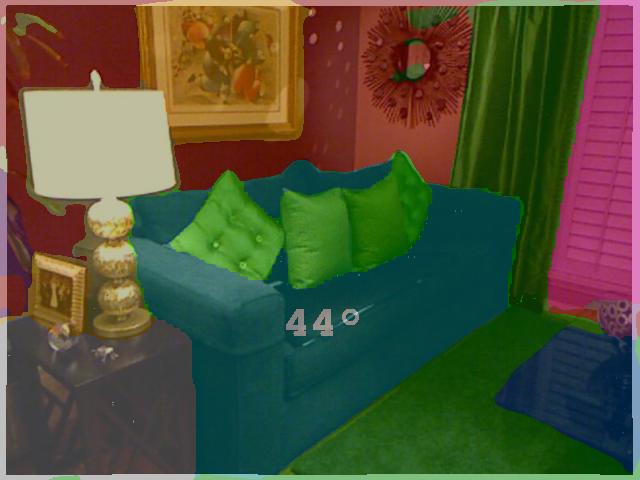}%
    	};%
    	\node[scene_label] at (18.03, -0.95){living room};%
    \end{tikzpicture}%
    \vspace{-4mm}%
    \caption{Qualitative results as RGB image overlayed with predicted panoptic segmentation, predicted scene class, and, for NYUv2, estimated orientations.}
    \label{fig:qualitative_results}
\end{figure*}

\subsection{Additional Pretraining on Hypersim}
\label{sec:experiments:pretraining_hypersim}
Finally, we examined how pretraining on the synthetic Hypersim dataset affects the performance of our derived multi-task setting for both NYUv2 and SUNRGB-D.
For further details on the pretraining, we refer to our implementation. 
The results are shown in Tab.~\ref{tab:results_nyuv2}~(MT~\uproman{5}) and Tab.~\ref{tab:results_sunrgbd}~(MT~\uproman{5}).
It turns out that, for NYUv2, especially mIoU and PQ greatly benefit from additional pretraining, while, for \mbox{SUNRGB-D}, only the performance of instance-related tasks is improved.
This can be deduced to the fact that SUNRGB-D alone is already much larger than NYUv2.
Fig.~\ref{fig:qualitative_results} shows qualitative results from pretraining on Hypersim and subsequent training on NYUv2. 
The latter represents our best network for NYUv2.

\subsection{Comparison to State of the Art}
\label{sec:experiments:comparison_to_sota}
Comparing our approach to the state of the art is challenging as tasks such as orientation estimation or scene classification have not been considered so far in related work due to missing annotations and a deviating class spectrum. 
Moreover, panoptic segmentation has not yet been attempted on NYUv2 or SUNRGB-D.
Therefore, we first established comprehensive single-task baselines~(see Fig.~\ref{fig:single_task}) covering common backbones ranging from sophisticated backbones to more efficient backbones that also enable mobile application.
Beyond that, we further trained well-known approaches for panoptic segmentation and scene classification on NYUv2, as shown in Tab.~\ref{tab:comparison}.
For Panoptic DeepLab, we applied the same parameters as described in Sec.~\ref{sec:main:panoptic_segmentation} for keypoint non-maximum suppression to postprocess instance centers.
To summarize, our proposed lightweight EMSANet achieves comparable or even better results than other approaches.
Moreover, larger backbones do not necessarily improve performance but significantly increase resource requirements.

\begin{table}[]
\vspace{-1mm}
\caption{
    Comparison to other state-of-the-art approaches on NYUv2 test split without test-time augmentation. %
    Legend: italic: metric used for determining the best checkpoint, *:~(re)training on our enriched NYUv2 dataset, $\dagger$:~ten-crop evaluation at 224$\times$224. pre.: additional pretraining on Hypersim.
}
\vspace{-2mm}
\resizebox{\linewidth}{!}{%
\scriptsize
\begin{tabular}{@{}l@{\hspace{2mm}}l@{\hspace{2mm}}l@{\hspace{4mm}}ccc@{}}
\toprule
                                                    & \textbf{Backbone}     & \textbf{Mod.}  & \textbf{mIoU}$\,{}^\uparrow$ & \textbf{PQ}$\,{}^\uparrow$    & \textbf{bAcc}$\,{}^\uparrow$  \\%
\midrule    
ESANet~\cite{esanet2021icra}                        & 2${\times}$R34-Nbt1d  & RGB-D          & 50.30                        & ---                           & ---                           \\%
ShapeConv~\cite{ShapeConv-cvpr2021}                 & ResNext101\_32x8d     & RGB-D          & 50.20                        &                               &                               \\%
\midrule
Panoptic DeepLab*~\cite{PanopticDeeplab-cvpr2020}   & R50                   & RGB            & 39.42                        & \emph{30.99}                  & ---                           \\%
                                                    & R101                  & RGB            & 42.55                        & \emph{35.32}                  & ---                           \\%
\midrule
MobileNetV2*~\cite{MobileNetv2-cvpr2018}            & $\mathrm{\alpha{=}1}$ & RGB            & ---                          & ---                           & 69.30$^\dagger$               \\%
EfficientNet*~\cite{EfficientNet-icml2019}          & B0                    & RGB            & ---                          & ---                           & 70.83$^\dagger$               \\%
Mod. ResNet34~(ours)                                & R34-Nbt1d             & RGB            & ---                          & ---                           & 70.25$^\dagger$               \\%
\midrule\midrule
\textbf{EMSANet (ours)}                             & R34-Nbt1d             & RGB            & 44.66                        & \emph{37.69}                  & 70.88                         \\%
                                                    & 2${\times}$R34-Nbt1d  & RGB-D          & 50.97                        & \emph{43.56}                  & 76.46                         \\%
                                                    & 2${\times}$R101-Nbt1d & RGB-D          & 50.83                        & \emph{45.12}                  & 77.41                         \\%
\midrule
\textbf{EMSANet pre. (ours) }                       & 2${\times}$R34-Nbt1d  & RGB-D          & 53.34                        & \emph{47.38}                  & 75.25                         \\%
\bottomrule
\end{tabular}
}
\label{tab:comparison}
\vspace{-5mm}
\end{table}
\section{Conclusion}
\label{sec:conclusion}
In this paper, we have proposed an efficient RGB-D multi-task approach for panoptic segmentation, instance orientation estimation, and scene classification, called EMSANet.
For training and evaluation, we have enriched the annotations of the common RGB-D indoor datasets NYUv2 and SUNRGB-D, which we also make publicly available.
To the best of our knowledge, we are the first to provide results in such a comprehensive multi-task setting for indoor scene analysis.
We have shown that all tasks can be solved using a single multi-task network.
Moreover, the individual tasks can benefit from each other when trained together.
Due to the efficient design, our approach enables fast inference, i.e. 24.5 FPS on an NVIDIA Jetson AGX Xavier and, thus, is well suited for mobile robotic applications.

\bibliographystyle{IEEEtran}
\bibliography{bib/literature.bib}

\end{document}